\documentclass[preprint,12pt]{elsarticle}

\usepackage{amssymb}
\usepackage{lipsum}
\usepackage{graphicx} 
\usepackage{subcaption}
\usepackage{hyperref}
\usepackage{booktabs}
\usepackage{xr}
\usepackage{xr-hyper}
\usepackage{hyperref}
\usepackage{xcolor}

\newlength{\subfigwidth}
\newlength{\subfigexamplewidth}
\newcommand{\expy}[1]{}
  
\usepackage{lineno}

\journal{Computers and Electronics in Agriculture}

\begin{document}

\begin{frontmatter}



\title{{H}olstein-{F}riesian {R}e-{I}dentification using {M}ultiple {C}ameras and {S}elf-{S}upervision on a {W}orking {F}arm}



\author[inst1]{Phoenix Yu\corref{cor1}}
\ead{ho19002@bristol.ac.uk}
\author[inst1]{Tilo Burghardt}
\author[inst2]{Andrew W Dowsey}
\author[inst1]{Neill W Campbell}

\affiliation[inst1]{organization={School of Computer Science},
            addressline={Merchant Venturers Building, Woodland Road, University of Bristol}, 
            city={Bristol},
            postcode={BS8 1UB}, 
            state={Bristol},
            country={United Kingdom}}


\affiliation[inst2]{organization={Bristol Veterinary School, University of Bristol},
            addressline={Langford House, Dolberry, Churchill}, 
            city={Bristol},
            postcode={BS40 5DU}, 
            state={Bristol},
            country={United Kingdom}}

\cortext[cor1]{Corresponding Author, PhD Student}


\begin{abstract}
We present MultiCamCows2024, a farm-scale image dataset filmed across multiple cameras for the biometric identification of individual Holstein-Friesian cattle exploiting their unique black and white coat-patterns. Captured by three ceiling-mounted visual sensors covering adjacent barn areas over seven days on a working dairy farm, the dataset comprises $101,329$ images of $90$~cows, plus underlying original CCTV footage. The dataset is provided with full computer vision recognition baselines, that is both a supervised and self-supervised learning framework for individual cow identification trained on cattle tracklets. We report a performance above \textbf{$96\%$} single image identification accuracy from the dataset and demonstrate that combining data from multiple cameras during learning enhances self-supervised identification. 
We show that our framework enables automatic cattle identification, barring only the simple human verification of tracklet integrity during data collection.
Crucially, our study highlights that multi-camera, supervised and self-supervised components in tandem not only deliver highly accurate individual cow identification, but also achieve this efficiently with no labelling of cattle identities by humans. We argue that this improvement in efficacy has practical implications for livestock management, behaviour analysis, and agricultural monitoring. For reproducibility and practical ease of use, we publish all key software and code including re-identification components and the species detector with this paper, available at \href{https://tinyurl.com/MultiCamCows2024}{https://tinyurl.com/MultiCamCows2024}.
\end{abstract}


\end{frontmatter}



\let\clearpage\relax
\section{Introduction}
\label{sec:introduction}
\textbf{Monitoring of Animals.} Today, sensor-based approaches for the study of animals including animal biometrics~\cite{kuehl2013} play a crucial role in various aspects of biological and veterinary research and practice~\cite{aguilar2023machine}. They provide a critical tool for the implementation of animal conservation and welfare policies~\cite{tuia2022,reynolds2024potential}. Vision-based Artificial Intelligence~(AI) methodologies, in particular, have become the technology of choice~\cite{pollock2025harnessing} for applications related to species conservation and biodiversity assessment~\cite{roy2023wildect,karaderi2024} as well as smart farming~\cite{li2021practices,andrew2020fusing}. In these settings, modern machine learning approaches often boost precision, efficiency, and flexibility of the monitoring tasks at hand. AI tasks range from species recognition, tracking and pose estimation, body condition scoring, behaviour detection and individual animal re-identification~(Re-ID).

\textbf{Motivation for Dairy Cattle Identification.} In agriculture specifically, keeping track of individuals in time and space by performing Re-ID tasks is a key requirement that underpins many applications ranging from welfare assessment to production management activities such as growth monitoring, estrus detection, and precision insemination timing. We note that Holstein-Frisian cattle are a particularly important species for which to focus efforts, as they form a foundation for dairy farming in many regions of the world. Moreover, the utilisation of 24/7 monitoring methods for this species promises wide-reaching improvements in both efficiency and also welfare in the dairy sector~\cite{akdeniz2022study}. However, expanded roll-out of such domain-specific AI systems first requires further improvements both in terms of a reduction in the amount of human input required for training, but also in the accuracy of system outputs. In any case, \textcolor{black}{self-supervised} cattle monitoring via Computer Vision~(CV) is a particularly promising avenue due to the remote operation of cameras without the need for physical sensor contact or reader stations positioned very close to the animals.

\textbf{Single Camera Systems.} Andrew et al.~\cite{andrew2017visual, andrew2019aerial, andrew2021visual} were one of the first to introduce automated CV frameworks for localisation and individual recognition of Holstein-Friesians based on their unique and (nearly) population-universal black and white coat patterns. To train these frameworks, datasets from single cameras were gathered from static~\cite{gao2021towards, gao2022label} and dynamic \cite{andrew2019aerial} observation platforms, then manually labelled, and finally used as information to facilitate deep machine learning for system construction. Yet, the required labelling work took weeks of manual labour and such a simple label-train-deploy strategy suffers from high re-commissioning cost, since species-generic features are repetitively retrained from scratch to allow for application to new herds or farms.

\textbf{Scenarios, Efficiency, and Transferability.} Using both individual and species-encompassing knowledge, based on large datasets followed by limited fine-tuning has been acknowledged as one efficient approach for the training and running of farm surveillance applications. To this end, Gao et.al.~\cite{gao2021towards}, for instance, created a relatively large single-camera dataset to enable experimentation regarding the minimisation of human-labelling for the identification of individual Holstein-Friesian cows from top-down CCTV. \textcolor{black}{Their work demonstrated the importance of self-supervision by reducing reliance on human labeling, establishing it as a strong backbone candidate for learning transferable representations to identify targets and guide future research.} However, using a single top-down camera limits identification to one view. It provides little knowledge regarding realistic monitoring performance for scenarios where multiple cameras are needed to cover a farm -- possibly with overlapping fields of view. \textcolor{black}{Consequently, customised re-training becomes unavoidable when applying a single-camera model across a multi-camera setup with varying angles, heights, and hardware configurations, significantly increasing human labour as the number and diversity of cameras grow.}

\begin{figure}[!htbp]
\centering
\captionsetup{justification=centering}
\includegraphics[width=15.5cm]{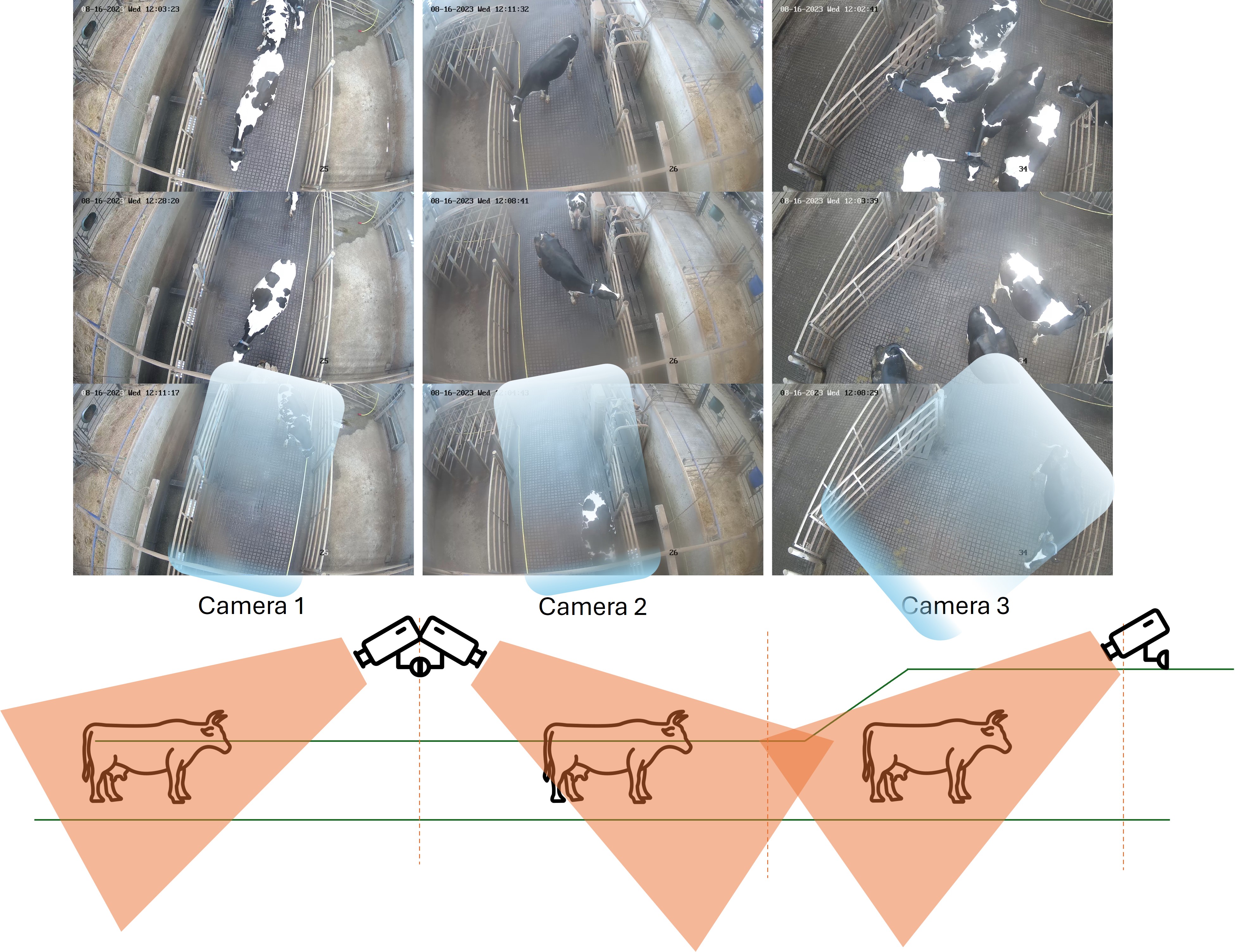}
\caption{\textbf{Multi-Camera Setup in our dataset.} The three views: cameras $1$, $2$, and $3$ from left to right, three different times vertically. The cameras are positioned so as to best capture cows following their regular milking routine. Cameras~$1$ and $2$ view a narrow milk-race which widens out in camera~$3$. In this wider area, cows are more likely to be stationary, resulting in an imbalanced image dataset. Additionally, the cow detection regions of interest are shown in the bottom row.}
\label{fig:perspective_example}
\end{figure}

\textbf{Objective: A Dataset for Multi-camera Monitoring.} Therefore, we introduce the first large, fully annotated, multi-camera dataset of Holstein-Friesian cattle with angled viewpoints taken on a working farm to study and develop a truly realistic and practically reusable monitoring scenario for dairy farms. Fig.~\ref{fig:perspective_example} illustrates the varying viewpoints of the system and the Graphical Abstract shows examples of cattle images as taken by the various cameras. We propose efficient baseline monitoring approaches with such a setup, report on the dataset's properties, and quantify results of deep learning approaches for automated re-identification of individual cattle for easy reproducibility and application. Firstly though, we will review the state-of-the-art in more detail and further explore why a dataset of the type introduced here is required to further the domain.

\section{Background}
\label{sec:background}
\subsection{Re-Identification and Farm Systems}
\textbf{Supervised Deep Learning for Static Domains}. Across domains, fully-supervised deep learning has achieved remarkable results regarding target categorisation and localisation,  and is the technology of choice for Re-Identification systems. Yet, its heavy reliance on fixed, annotated data at training time does not deal well with some common problems, such as the introduction of new animals with novel patterns. Training large classifiers from scratch under such settings is inefficient and often not feasible in practical, economically constrained settings.

\textbf{Learning in Dynamic Data Domains}. Adaptive transfer learning and self-supervised learning have aided address such dataset-wide scenarios by transferring meaningful knowledge from an existing model onto unlabelled datasets that retain some structure (e.g. the same animal species -- even if patterned completely differently). Reliant on a ResNet backbone and spectral embedding analysis and able to propose accurate identification over smaller batches of data efficiently, Pathak et al.~\cite{pathak2020fine} refined earlier human Re-ID tools through inner fine-graining across spatial subregions of latent space. Therien et al.~\cite{therien2023towards} then extended this Re-ID framework for 3D depth sensor applications by utilising PointNet and ViT. We note that these approaches are, in principle, fully applicable to the Holstein-Friesian cattle Re-ID task.

\textbf{State-of-the-Art for Cattle Re-Identification Systems}. Sharma et al.~\cite{sharma2024universal} utilise depth data for universal bovine identification from top-down images, while Bhole et al.~\cite{bhole2019computer} introduced a pipeline fusing standard RGB and thermal images to monitor individual cows. Recent work by Dubourvieux et al.~\cite{dubourvieux2023cumulative} involved a Cumulative Unsupervised Multi-Domain Adaptation~(CUMDA) strategy for Re-ID tasks over multiple coat pattern-centred cow datasets~\cite{bhole2019computer, gao2021towards} with varied perspectives, lighting, and occlusion. Their work suggests that unsupervised strategies may have an advantage over self-supervision since the latter contained a higher cost for pre-labelling diversely conditioned images. However, given mainly torso information, Gao et al.~\cite{gao2022label} exploited a self-supervision strategy for fast Re-ID built on a top-down view dataset of Holstein-Friesian cows only. We hypothesise that self-supervision strategies and new datasets can be exploited further for transferring and generalising species-specific identification knowledge across varied visual conditions -- albeit requiring the construction of a true multi-camera dataset.

\subsection{Species Recognition}
\textbf{CNN-based Object Recognition.} Deep residual CNNs and their derivatives are the most widespread toolkits for semantic detection and segmentation. The introduction of region proposal networks~(RPNs) allowed for performing such tasks faster and perform real-time segmentation tasks. Later, pyramid-shaped networks utilised stacking region selection with inter-layer knowledge to acquire multi-level features. Knowledge of the previously trained semantic segmentation model can also be further distilled for transfer learning over advanced and detailed features. Based on these and related approaches, animal recognition and even behaviour analysis has not only been applied to wild species~\cite{sakib2020visual}, but also to Holstein-Friesian cattle~\cite{nguyen2021video} including work by Andrew et.al.~\cite{andrew2019aerial} proposing a drone-based, automatic CNN framework to monitor Holstein-Friesian cows from the air.

\textbf{Transformer-based Object Recognition.} Precise segmentation performance in higher-resolution images is often reduced when using a traditional CNN architecture due to the fixed-sized operations within blocks. An alternative, first used in language models, introduces various versions of Vision Transformers (ViT) to the task of object recognition. However, ViTs and similar architectures face challenges regarding effective token sampling and correlation acquisition. In response, Fayyaz et.al.~\cite{fayyaz2021ats} boosted ViT performance with an adaptive token sampling module and Yu et.al.~\cite{yu2022metaformer} replaced attention layers with a pooling module and built a lightweight model on top of a general MetaFormer architecture.

\textbf{Reliable Species Detection.} Expanding on earlier, pioneering work by Andrew et al.~\cite{andrew2017visual, andrew2019aerial, andrew2021visual} and others, cattle species detection is a widely solved task where both CNN and Transformer solutions enable reliable detection ~\cite{sharma2024universal, wang2024ultra}. Potential detection errors may still occur in domains with high levels of occlusion, low image quality or highly unusual viewpoints. However, whilst detection can utilise datasets of the species agnostic to individual appearance, individual animal re-identification is by no means a fully solved challenge and approaches must pay close attention to individual variability across viewpoints that can cover a whole farm -- these properties are largely inaccessible in previous datasets related to re-ID.

\subsection{Individual Identification}
\textbf{Construction of Latent Spaces for Individuals.} In order to classify a population, a network is usually tasked with mapping the visual appearance of individuals into a distinctly clustered latent space, where clusters relate to individuals. Contrastive learning can be used to shape such a space and metric learning~\cite{andrew2021visual}, in particular, aims at building the space such that distances in latent space reflect the relationships between the instances of the population. Chen et.al.~\cite{chen2020simple} built SimCLR in which faster and more accurate results can be learned via a selection of prior data augmentation. \textcolor{black}{They utilised Normalized Temperature-scaled Cross Entropy Loss~(NTXentLoss) to maximise the distance between all unique labels.} We utilise metric learning for clustering the individually-indicative latent space building on work by Gao et al.~\cite{gao2022label}. Before discussing these experiments we will first outline how the core dataset was collected and prepared.

\section{Methodology}
\label{sec:methodology}
\subsection{Dataset Acquisition}

\textbf{Filming on a Working Farm.} Videos of Holstein-Friesian cows were acquired daily, shortly after each animal departed an automatic milking system. Three spatially adjacent cameras positioned along a narrow walkway~(see Fig.~\ref{fig:perspective_example}) were used. All videos were acquired during the lunchtime milking session (12:00 PM - 2:00 PM) using HikVision DS-2CD5546G0-IZHS IP cameras running at a framerate of 25FPS and a resolution of $2560 \times 1440$ pixels per frame. The videos from each camera were encoded as MP4 files on-device and downloaded remotely for processing.

\subsection{Tracklet Generation and Refinement}

\begin{figure}[!ht]
\centering
\captionsetup{justification=centering}
\includegraphics[width=12.5cm]{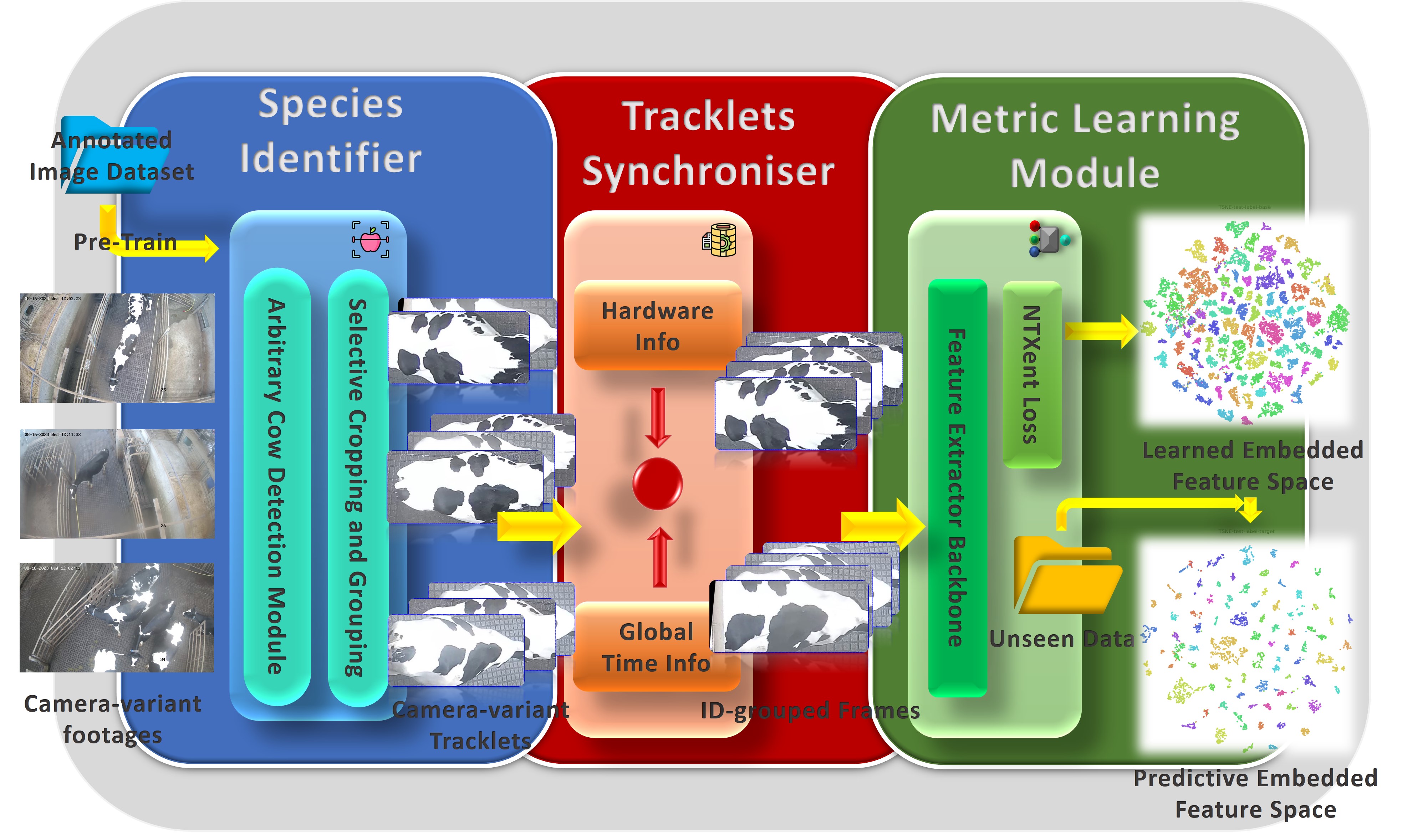}
\caption{\textbf{Processing Pipeline Overview.} Our framework consists of three parts. First, we train and apply the species identifier~\textbf{(left)} to acquire tracklets of cows from the video series. Then, we merge tracklets of cows over cameras. Human inspections at this stage and guarantees tracklet correctness and integrity~\textbf{(middle)}. Finally, we train and evaluate our metric learning modules~\textbf{(right)} for individual Re-ID on both supervised and self-supervised pipelines. Note that no manual labelling of cattle IDs is required in the latter.}
\label{fig:Full_map}
\end{figure}
\begin{figure}[!ht]
\centering
\captionsetup{justification=centering}
\includegraphics[width=12.5cm]{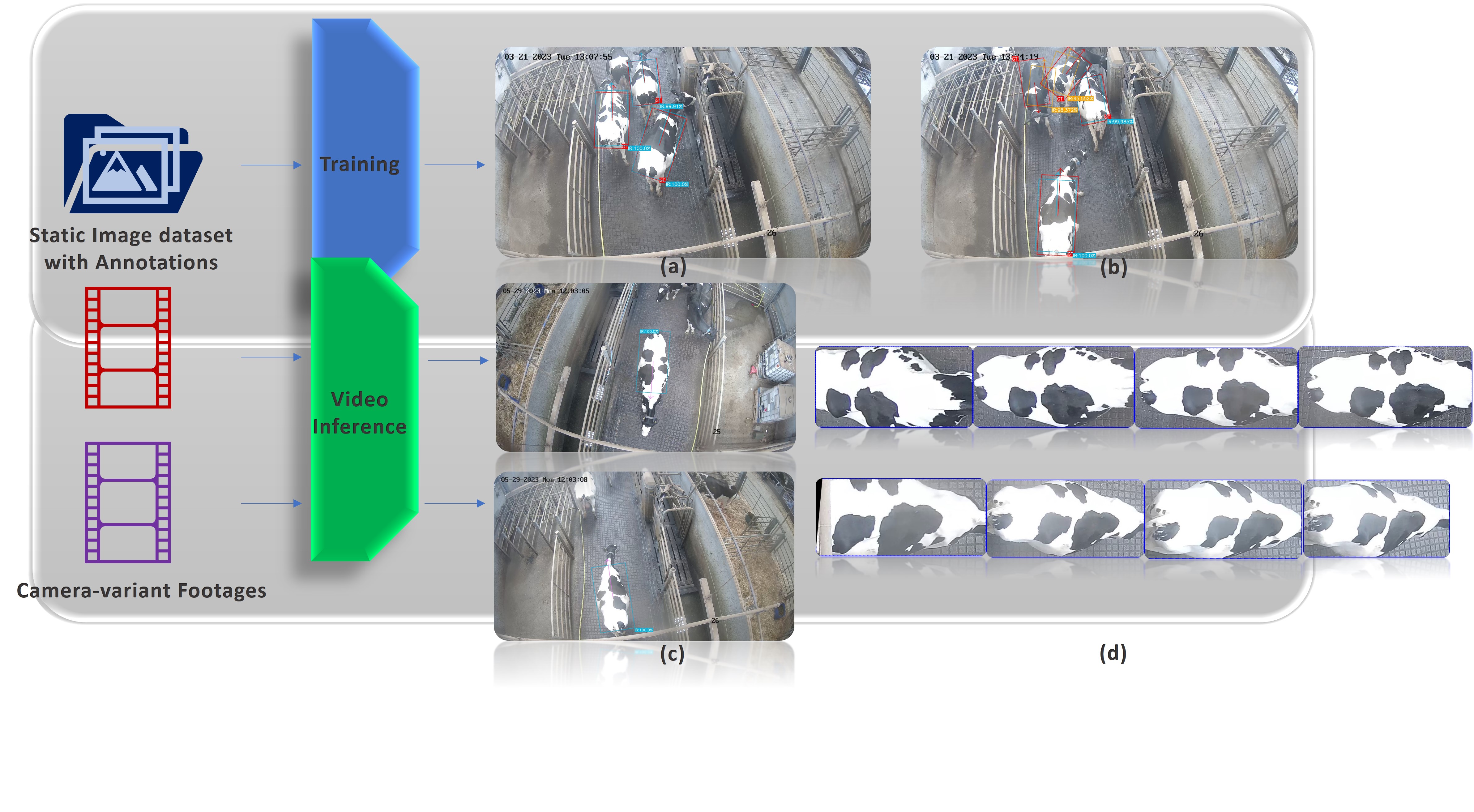}\vspace{-23pt}
\caption{\label{fig:SI_map}\textbf{Species Identification.} An overview of our Species Identifier. Initially~\textbf{(upper)}, we train an arbitrary cow detector using image datasets with COCO format annotations. \textbf{(a)} represents a good image inference~(\textit{blue}) compared to ground-truth~(\textit{red}), whereas \textbf{(b)} contains false positives~(\textit{yellow}). After cycles of tuning and retraining, we then apply the trained model~\textbf{(lower)} on videos for each of the cameras. \textbf{(c)} is an example of detections from two different cameras. Cattle are grouped to eventually generate camera-variant tracklets~\textbf{(d)} of cow torsos.}
\end{figure}

\begin{figure}[!htbp]
\centering
\captionsetup{justification=centering}
\begin{subfigure}{0.49\textwidth}
\centering
\includegraphics[width=\textwidth]{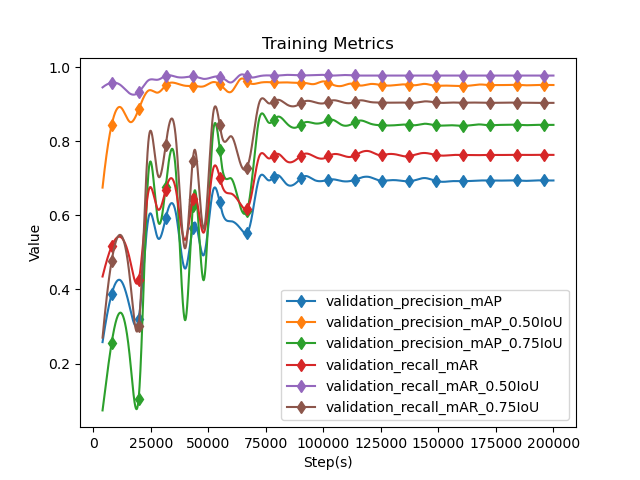}
\caption{mAP and mAR during training/validation}
\label{fig:metric}
\end{subfigure}
\begin{subfigure}{0.49\textwidth}
\centering
\includegraphics[width=\textwidth]{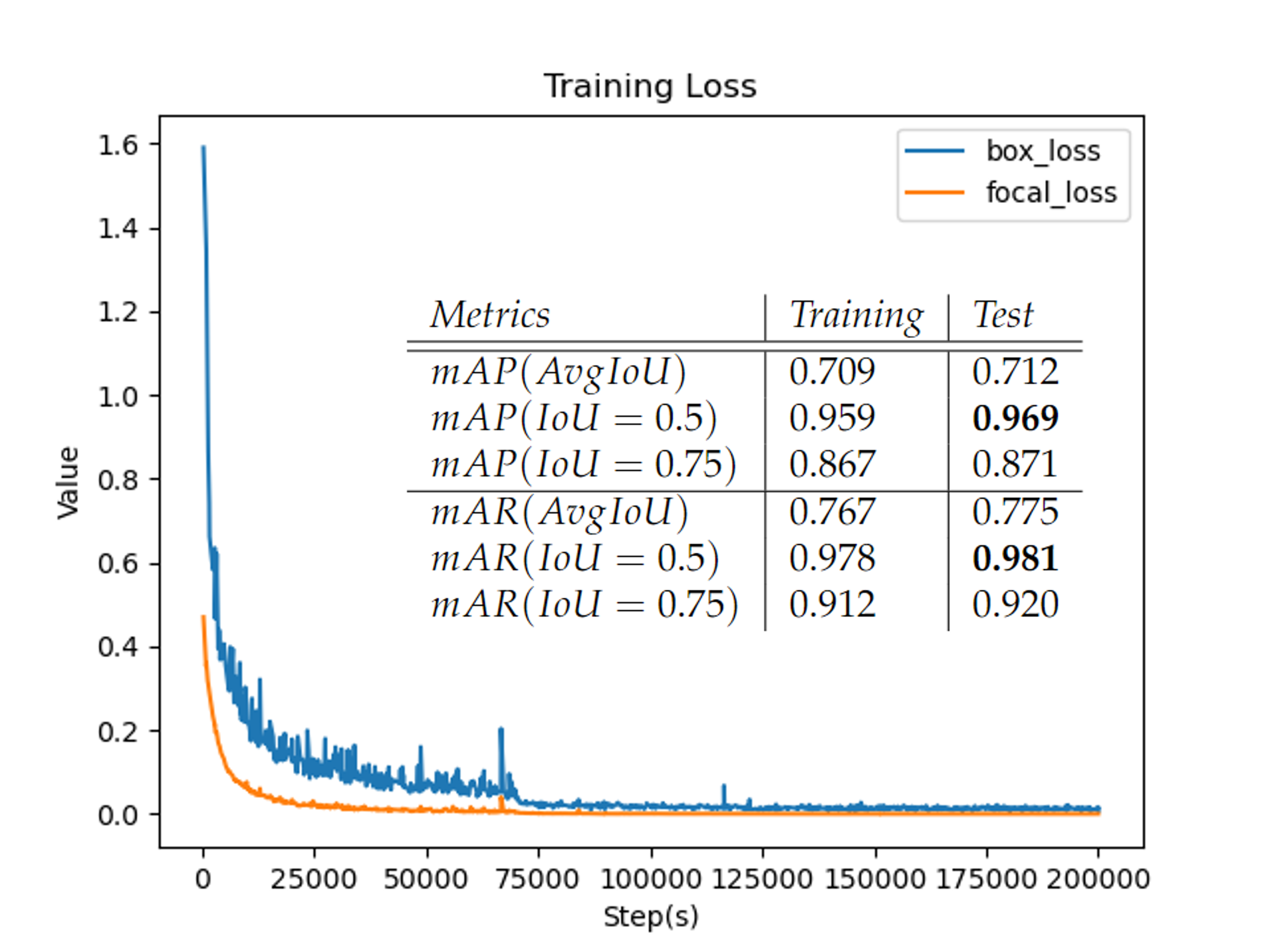}
\caption{Focal loss and Box loss}
\label{fig:losses}
\end{subfigure}\\[1ex]
\hfill
\caption{\textbf{Species Detector Training and Performance.} Training metrics~(mean average precision~(mAP) and mean average recall(mAR))~\ref{fig:metric} and losses~\ref{fig:losses} for our species detector, with a detailed table of test set metrics.}
\label{fig:all_metrics_si}
\end{figure}

\textbf{Species Detection and Tracklet Generation.} \textcolor{black}{Adapted from previous works by Gao et.al.\cite{gao2021towards}}, we first applied animal detection~\cite{andrew2021visual} to the farm footage in order to extract cattle locations for each frame in the CCTV video~(see Fig.~\ref{fig:SI_map}). We trained this species detector for cow torso recognition using pre-trained weights from ImageNet for ResNet-152 as our backbone. We adapted the tools from NVIDIA Computer Vision Toolkit\footnote{https://github.com/NVIDIA/retinanet-examples} to perform head-oriented, rotated bounding box detections, which are later normalised so that all images to have the same orientation -- i.e. the cow is always seen walking left-to-right in the dataset. Due to the wide visual range of views, we limited detection to a region of interest~(RoI) in the cattle lane for each camera~(see Fig.~\ref{fig:perspective_example}). Finally, we fine-tuned our species detector via Adam optimiser using $5,606$ annotated frames across three selected views. Fig.~\ref{fig:all_metrics_si} shows a training and performance overview for this species detector.

\textbf{Tracklet Generation.} Bounding box information of cow torsos are next collected into raw tracklets covering individual cattle passing the cameras. We sample the CCTV video at $5Hz$ to generate frames. Simple heuristics regarding cow movement and changes of direction are used to automatically concatenate detections from each consecutive frame into tracklets. In practice, we assume that cows cannot move more than $120$ pixels and rotate drastically beyond $45^{^\circ}$ between contiguous frames. Additionally, we manually checked the integrity of tracklets and removed segmentations with less than 80$\%$ of a valid torso or those containing overlapping crops due to multiple cows standing too closely together. Furthermore, we manually refitted crops affected by proximity errors {\it ad hoc} to maximise tracklet quality in the dataset. As we assume one tracklet per cow per day, the results of cleaned tracklets across seven days have labels referring to the same set of cows. Therefore, daily merging of identification (ID) tags was carried out based on the first day for each cow to enable later learning. Note that the final version of tracklets in the dataset still contains imbalanced frame quantities~(see Appendix Fig.~\ref{fig:metadata_hist}) and sizes~(Fig.~\ref{fig:metadata_sizes} for full details) across views. It can be seen from the floor plan~(Fig.~\ref{fig:perspective_example}) that each camera has disparate RoIs. Cows from both cameras~1 and 2 walk in a straight line in the middle, whereas camera~3, having several exits, leads to many more stationary cows waiting to leave the facility. \textcolor{black}{Ultimately, we have merged the raw, separate tracklets of individuals into a smaller set of refined tracklets, each representing progress of a cow across all three cameras.} Within a given camera, bounding box sizes of moving cows contain variations due to different camera elevations and the inherent perspective effect as cows get closer to the camera. The average size of our tracklet crops is $568 \times 264$, and more information on dataset metadata can be found in the Appendix.

\begin{figure}[!htbp]
\centering
\captionsetup{justification=centering}
\begin{subfigure}{0.49\textwidth}
\includegraphics[width=\textwidth]{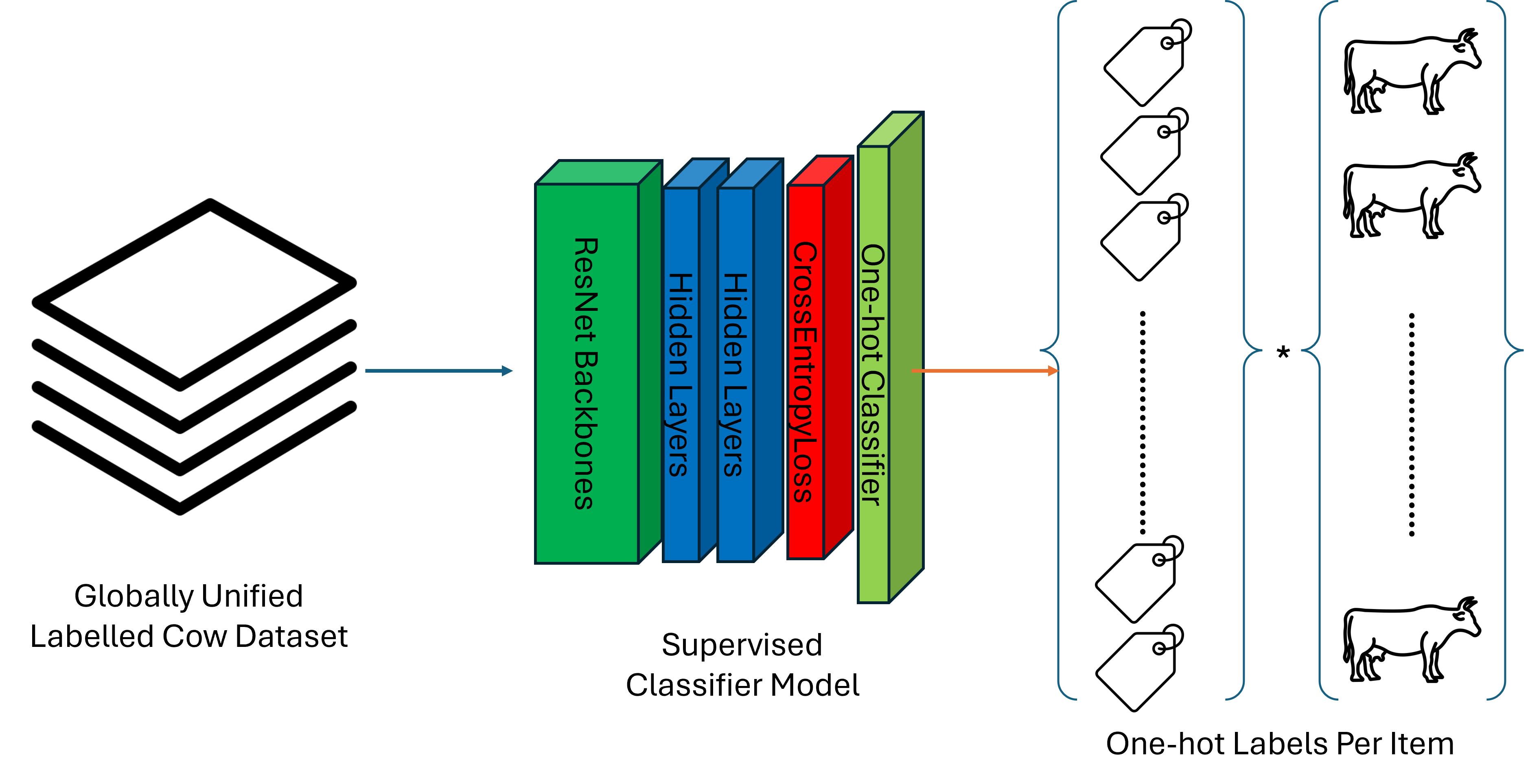}
\caption{}
\label{fig:super_workflow}
\end{subfigure}
\hfill
\begin{subfigure}{0.49\textwidth}
\includegraphics[width=\textwidth]{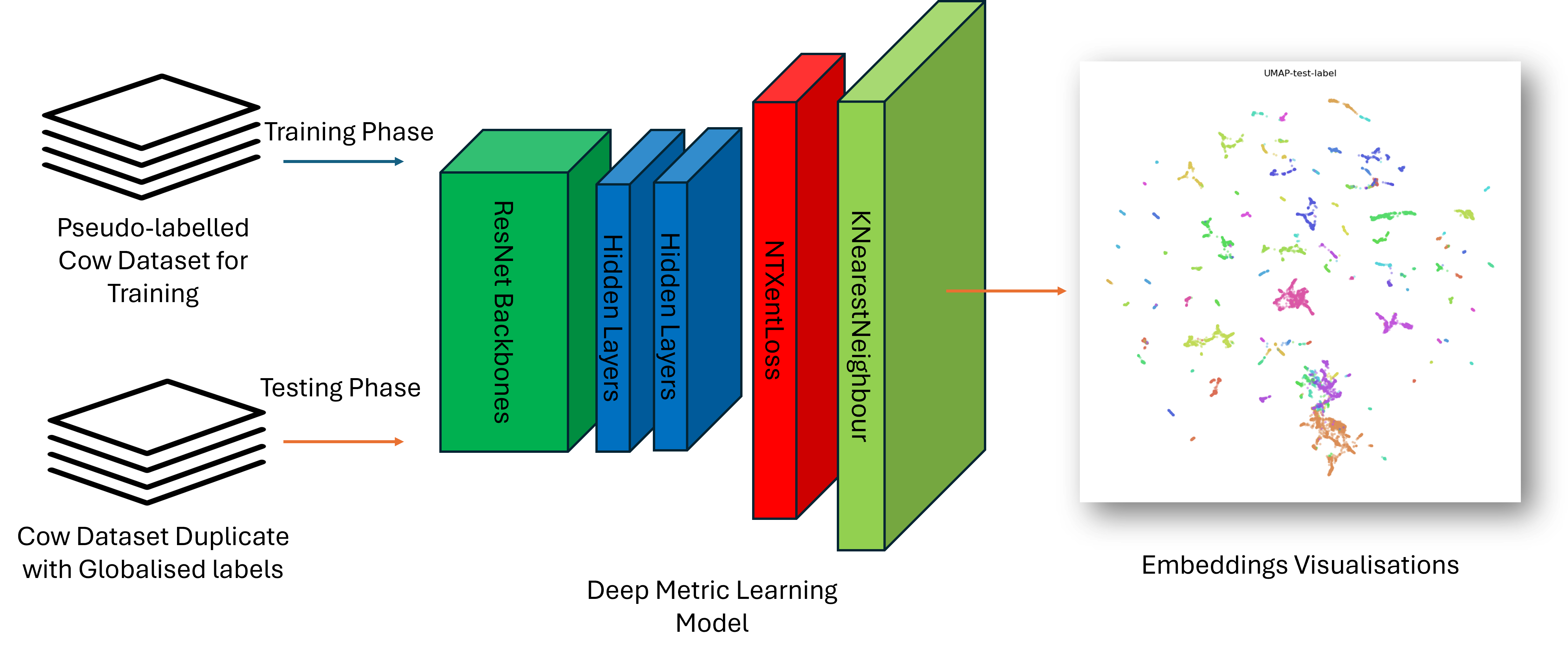}
\caption{}
\label{fig:self_workflow}
\end{subfigure}
\caption{\textbf{Deep Metric Learning.} Schematic Overview of our metric learning modules. Supervised~\textbf{(a)} adapts a classifier structure for predicting correct cow labels~(IDs). Self-supervised~\textbf{(b)} firstly trains with pseudo-labels (tracklet number) via metric learning. Then this model is used with the ground-truth~(IDs) for evaluation and embedding visualisation.}
\label{fig:ml_workflow}
\end{figure}

\subsection{Cattle Re-Identification}
\textbf{Latent Space Construction.} After generating uniformly-labelled tracklets which now reflect the 90 total cows we have collected features from, we proceeded to conduct supervised~(Fig.~\ref{fig:ml_workflow}-(a)) and self-supervised~(Fig.~\ref{fig:ml_workflow}-(b)) learning. Subsequently, we evaluate performance. Both identification approaches employed ImageNet pre-trained ResNet18 backbones responsible for embedding images into feature vectors of the latent space. Before inputting images into the models, we normalize
them to have a uniform size and photometric variance. Our training dataset comprises tracklets from the initial five days of data. Day six was used for validation -- and
day seven as the test set. Training was conducted via the AdamW optimiser at a learning rate of $0.001$ sampled at $201$ epochs~(chosen due to minimal validation loss at this point) using an NVIDIA RTX2080Ti system with 48GB of GPU memory. Training ran for $2873$ minutes overall. The full training process is illustrated in~Fig.~\ref{fig:all_metrics_si}~(left).

\textbf{Loss Function, Evaluation, and Experimental Setup.} Within the supervised framework, we employ a model incorporating cross-entropy loss and a one-hot classifier on top of the reshaped images. To address variations in quantity amongst individuals~(i.e. class imbalances), we sample images of individuals with equal probability for training.
Within the self-supervised framework, a linear layer was utilised to generate multi-dimensional feature vectors for metric learning purposes. \textcolor{black}{The NTXentLoss~\cite{chen2020simple} was used as a contrastive loss function. It encourages distance reduction of items having the same label in latent space while moving these items away from those with contrasting labels.} To ensure that the ground-truth identities of the cows could not `leak' into the
self-supervision process, pseudo-labels were used (effectively randomizing the order in which cows appeared daily).
For testing our unsupervised classifier, the ground-truth labels were used and KNN clustering was applied to evaluate performance. \textcolor{black}{Both the supervised and self-supervised frameworks have been trained using NVIDIA GeForce RTX 2080 Ti GPUs.}


\section{Results}
\label{sec:results}
\begin{figure}[!htbp]
\centering
\captionsetup{justification=centering}
\begin{tabular}{p{0.49\textwidth} p{0.49\textwidth}}
\begin{subfigure}[b]{0.49\textwidth}
    \includegraphics[width=\linewidth]{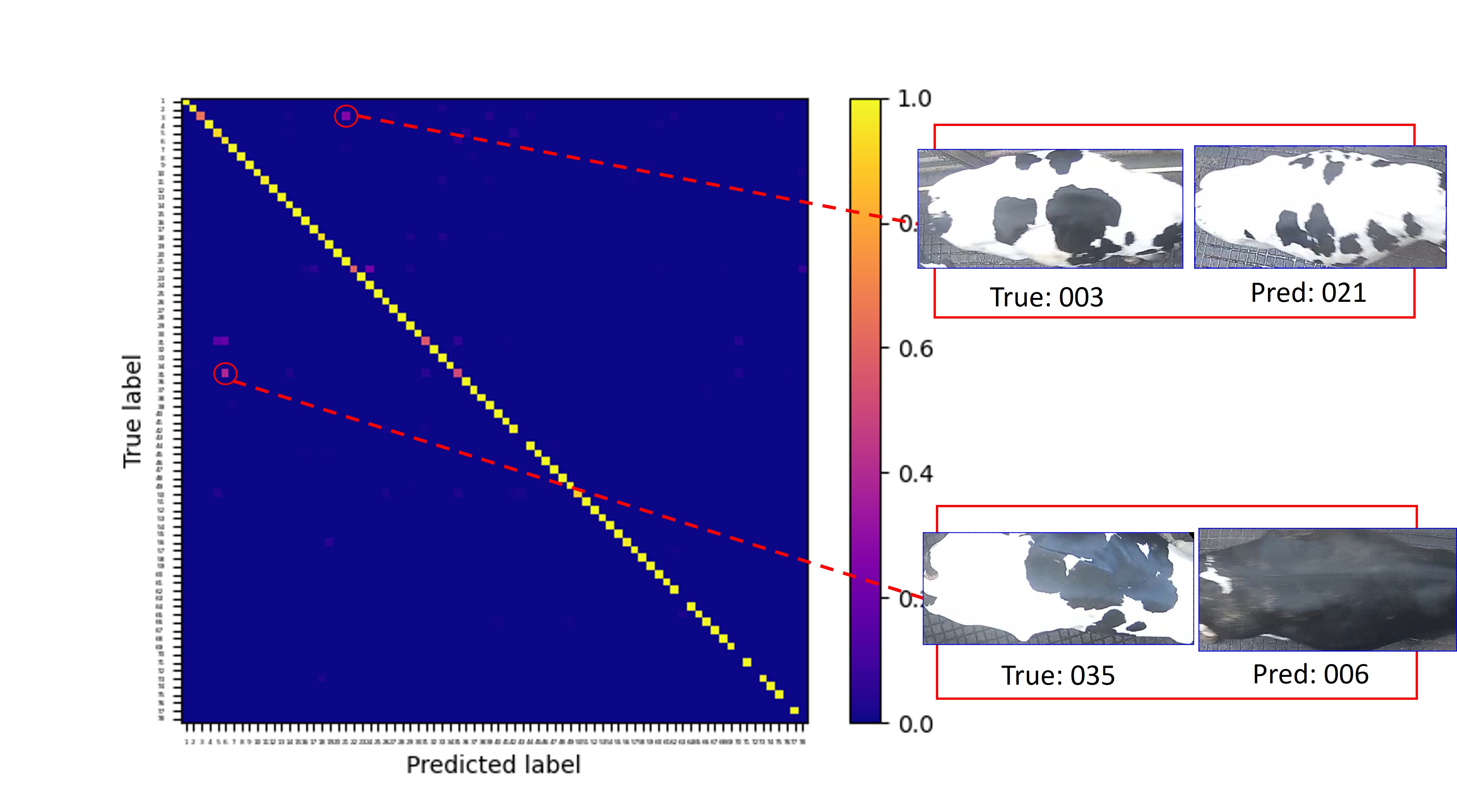}
    \caption{Fused}
\end{subfigure} &
\begin{subfigure}[b]{0.49\textwidth}
    \includegraphics[width=\linewidth]{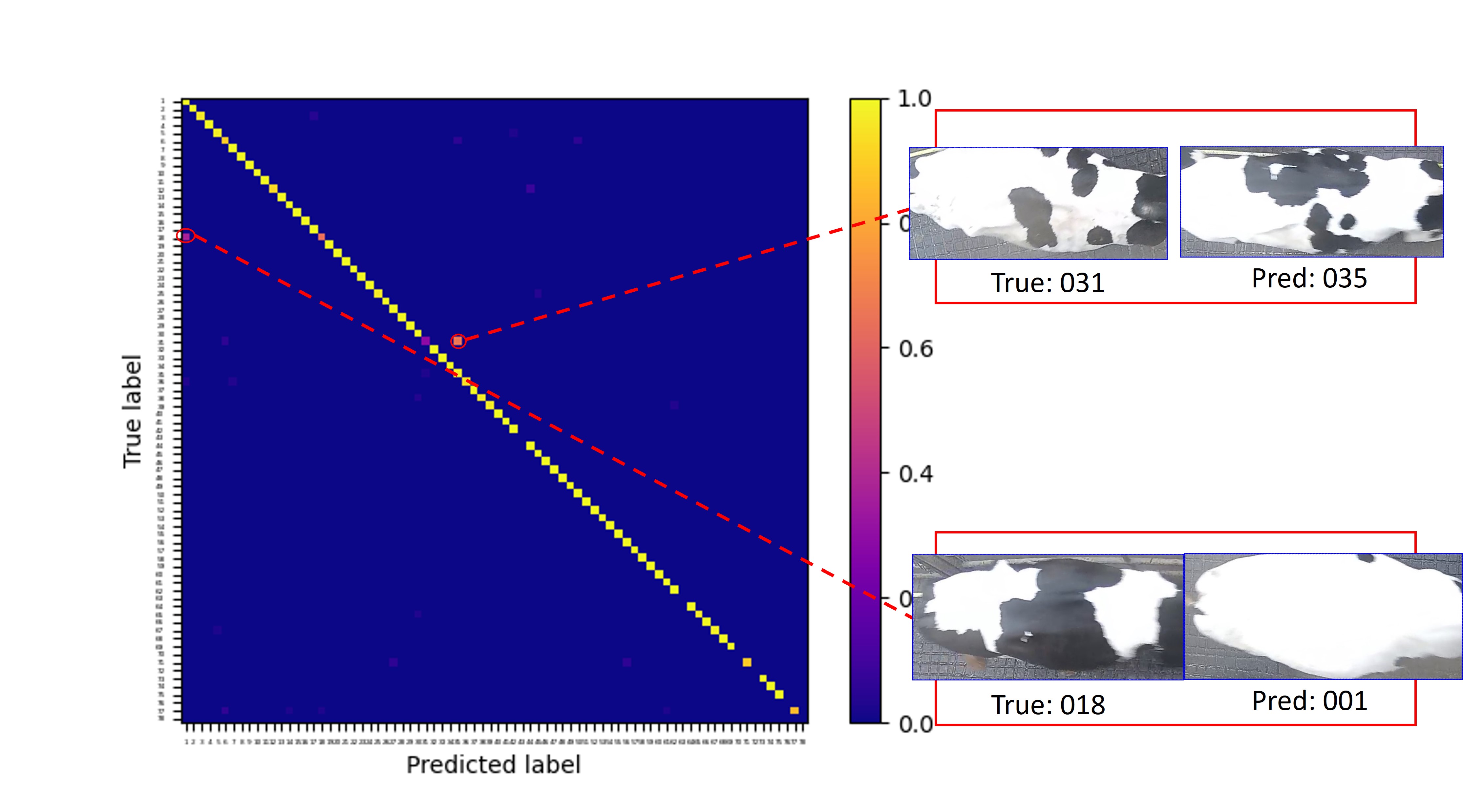}
    \caption{Camera 1}
\end{subfigure} \\

\begin{subfigure}[b]{0.49\textwidth}
    \includegraphics[width=\linewidth]{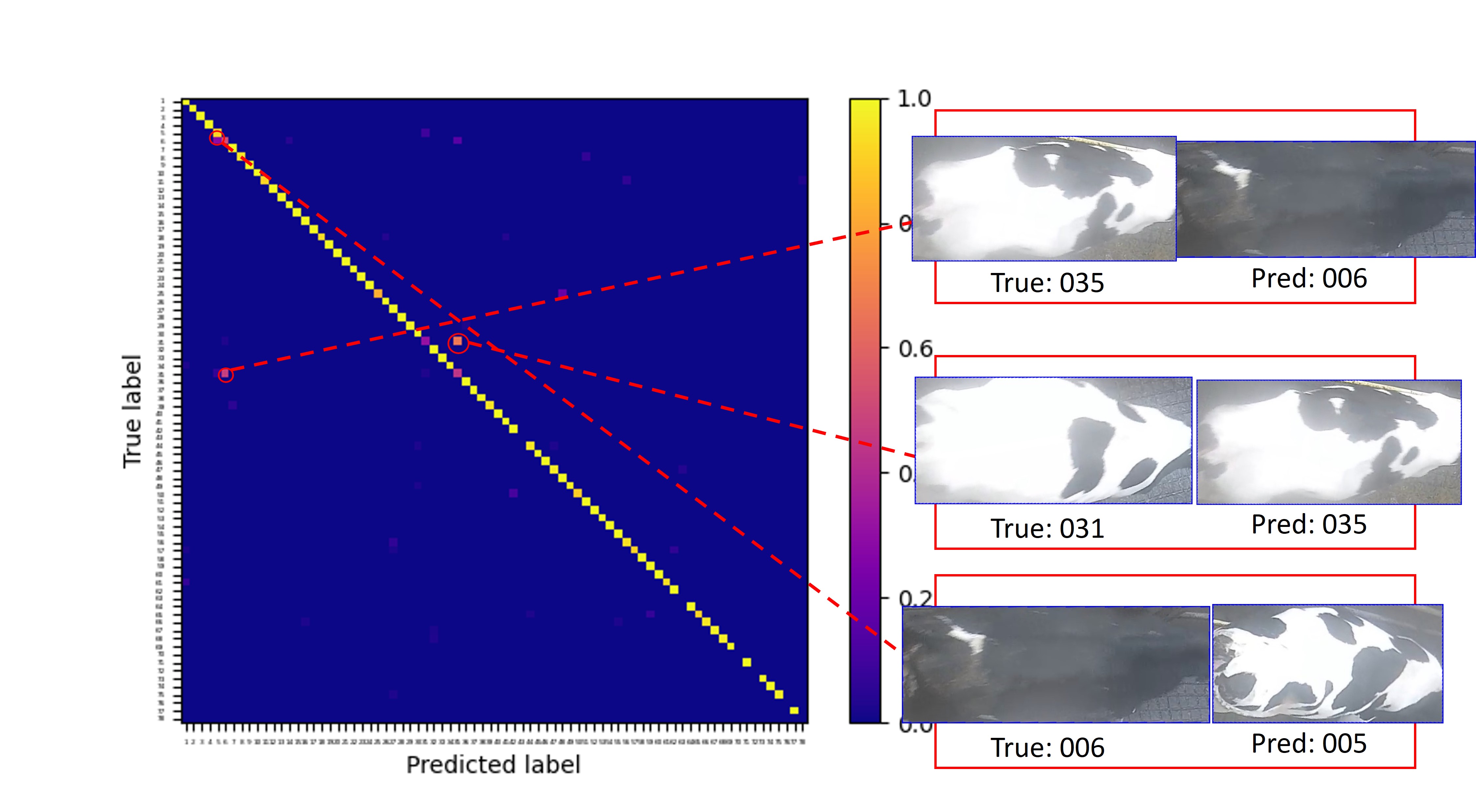}
    \caption{Camera 2}
\end{subfigure} &
\begin{subfigure}[b]{0.49\textwidth}
    \includegraphics[width=\linewidth]{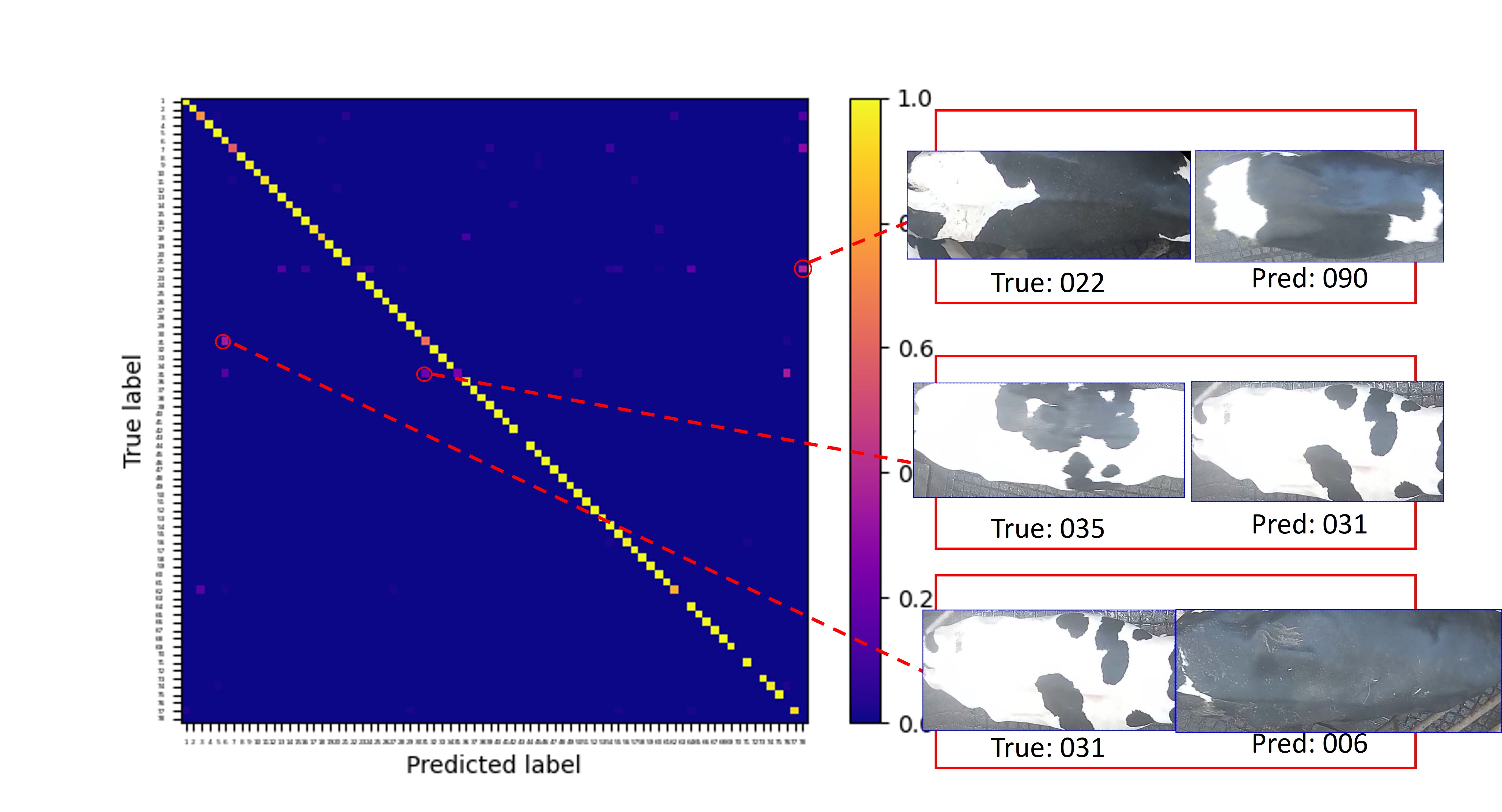}
    \caption{Camera 3}
\end{subfigure}
\end{tabular}
\caption{\textbf{Confusion matrices.} Confusion matrices of ground truth (true) versus prediction (pred) labels from the fused and all three singular cameras trained with the ResNet18 backbone, image size of $128$ and a hidden dimension of $64$. For each instance, examples of false positives are highlighted and shown with corresponding image samples on the right side. Note that cow No.035 is often misinterpreted as cow No.031. As seen from the image crops, they share an almost identical body pattern, with their rears being completely white and their lower side having black clusters of similar position and orientation.}
\label{fig:confusion_matrices}
\end{figure}

\begin{figure}[!htbp]
\centering
\captionsetup{justification=centering}
\begin{subfigure}{0.49\textwidth}
\includegraphics[width=\textwidth]{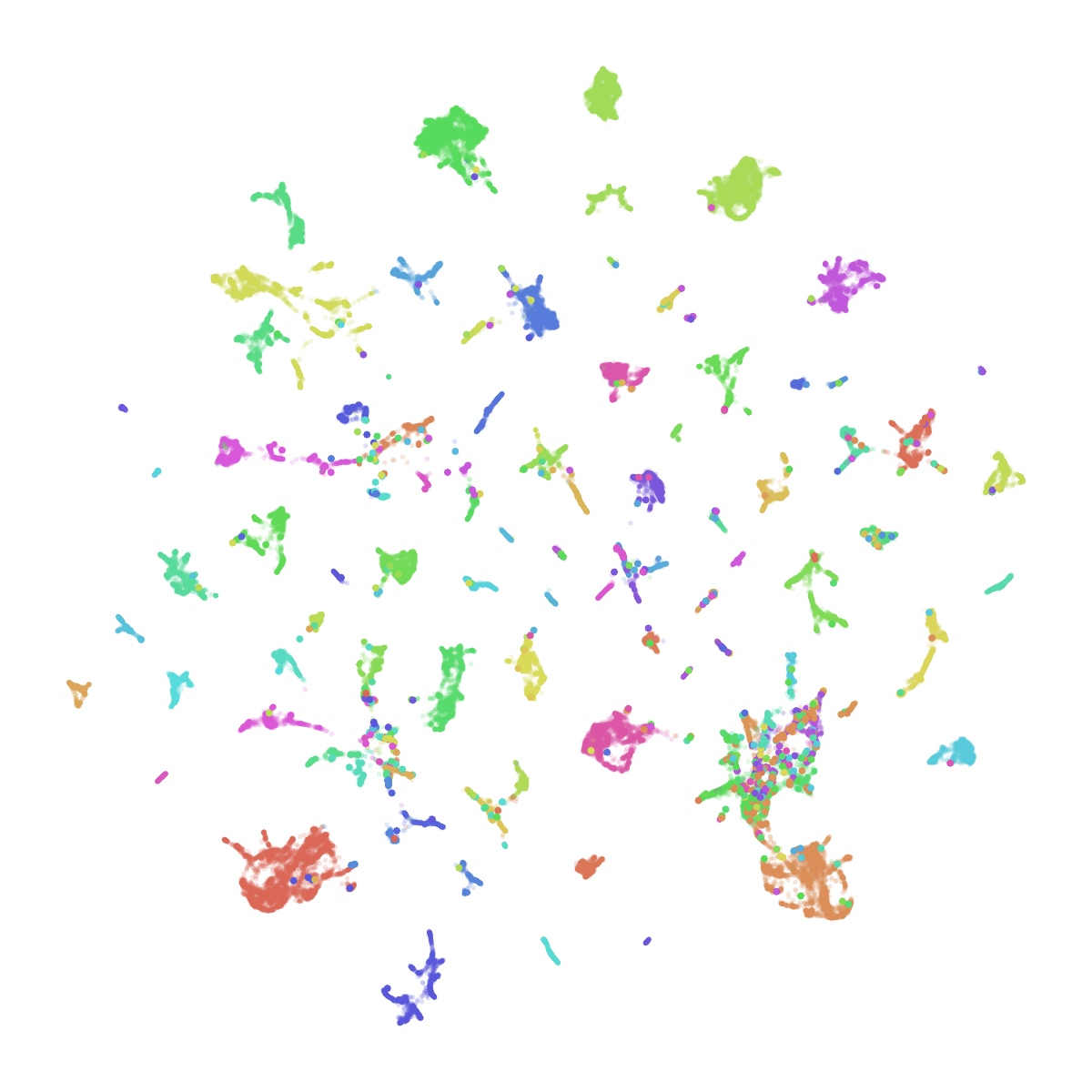}
\caption{Training data projected on trained embeddings}
\end{subfigure}
\hfill
\begin{subfigure}{0.49\textwidth}
\includegraphics[width=\textwidth]{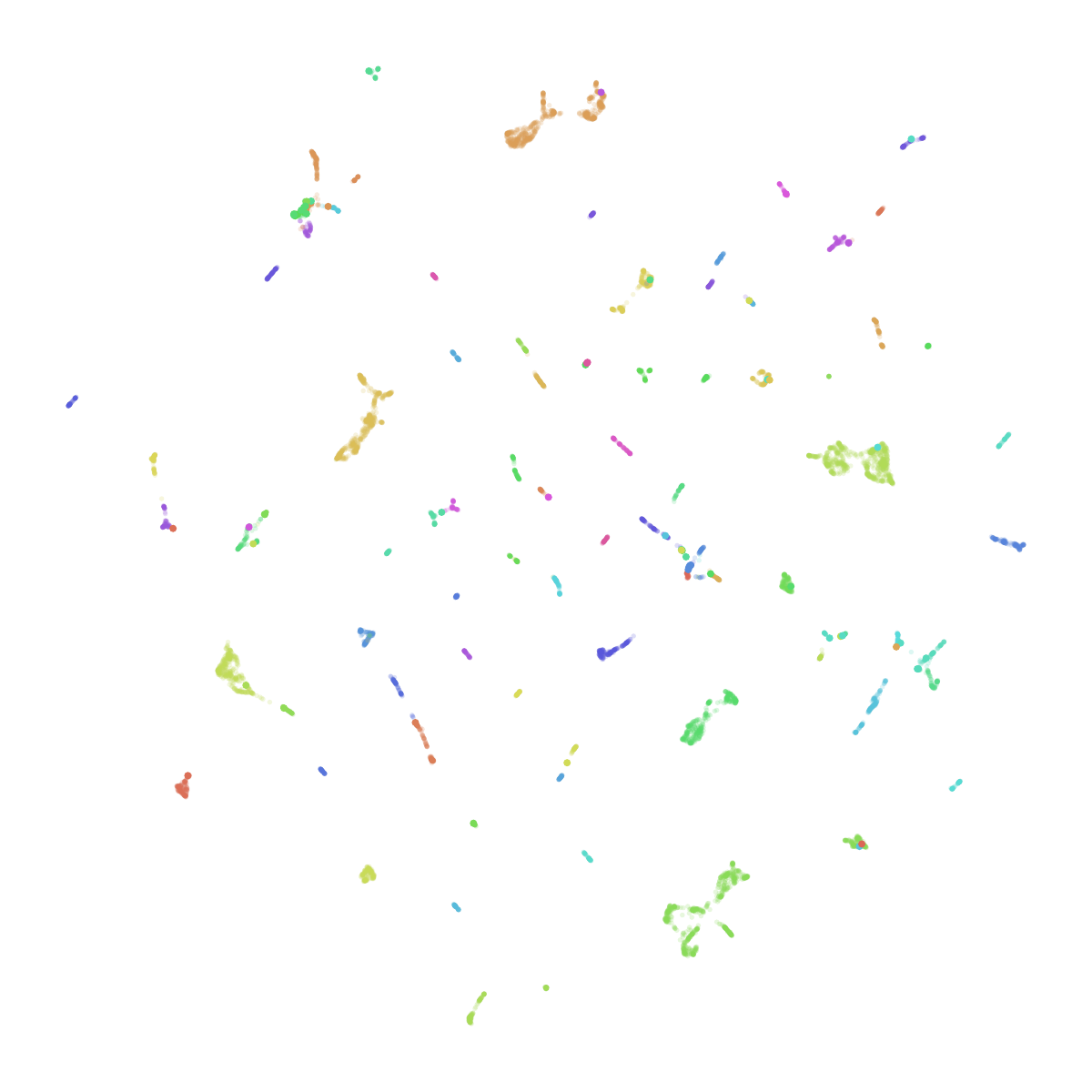}
\caption{Inference data projected on trained embeddings}
\end{subfigure}
\caption{\textbf{Self-supervised embedding visualisations.} Visualisations on feature embeddings based on UMAP. Row~\textbf{(left)} visualises KNN from training embeddings alone, while~\textbf{(right)} projects test data into the embedding space constructed on the training data.}
\label{fig:umap}
\end{figure}

\textbf{Re-Identification Performance.} Here we present results on the task of individual identification based on training from tracklets as described above. Table~\ref{tab:full_table} represents the quantitative performance of our supervised classifier and self-supervised identifier, respectively, where we compared the results of training for each independent camera and also with them all combined. For our classifier trained via supervised methods, the performance of merging the camera views reached the best accuracy result at $96.06\%$ with the cropped image size being $128 \times 128$ using a $32$D-embedding. This is marginally inferior to camera~$1$ performance ($96.78\%$), but is significantly improved compared to previously published single camera performance in slightly different settings benchmarked at $76.9\%$ accuracy as reported for Cows2021~\cite{gao2021towards}. Self-supervision demonstrated a small overall increase in performance when merging multiple views together as shown in Table~\ref{tab:full_table}. When training with our ResNet18 backbone with images resized to $128 \times 128$, combining views leads to an accuracy of $96.36\%$, boosting the performance of camera~$1$, $2$ and $3$ by $0.48\%$, $3.62\%$, and $39.67\%$ respectively. \textcolor{black}{With more images causing dataset imbalance}, the relatively large gap between the performance of camera $3$ versus the combined views was due to the cows in camera $3$ being mostly stationary (they tend to stop walking after the milk-race, so group together making it difficult to disambiguate individuals accurately). Cattle detection in dense groups  (due to a camouflaging effect of the coat pattern) is a truly difficult task not solved to the same quality levels as standard object detection. Fig.~\ref{fig:umap} and~\ref{Visualisations} provide visualisations of embeddings obtained during inference from self-supervision. Additionally, in Fig.~\ref{fig:confusion_matrices} we show the confusion matrices of ground truth vs.\ prediction. Note that pattern similarity remains the major reason for misidentification. For instance, cow $035$ is often misclassified as cow $031$ due to the similarity of their body patterning from the rear and the side.

\textcolor{black}{\textbf{K-Fold Cross Validation.} In addition, we applied K-Fold Cross Validation~(KFCV) to fully evaluate model test performances in both fully- and self-supervised settings(Table~\ref{tab:kfcv_table}). We chose the standard $k=10$ for cross validation after shuffling all the data for our fully-supervised pipeline. Since data formatting and training logic are significantly different for our self-supervised pipeline where cows were categorised in dates before in ID labels, we chose a customised $k=7$ based on 7 days from our data source for cross-validation. To avoid deviating from our training details, we sequentially cycle through each of 7 days as the test set and perform training and validation from the remaining 6 days, with 5 days as the training set and one as validation.}

\begin{table}[!htbp]
    \centering
    \begin{tabular}{c|c|c|c}
    \toprule
          &       & \multicolumn{1}{c}{\textbf{Supervised}} & \multicolumn{1}{c}{\textbf{Self-Supervised}} \\
    \midrule
    \textit{Dataset} & \textit{Hyperparameters} & \textit{Inference($\%$)} & \textit{Inference($\%$)} \\ \hline
    $All Cameras$ & Res18; img128; hid32 & 96.06 & \textbf{\underline{96.36}}\\
    $Camera 1$ & Res18; img128; hid32 & \textbf{96.78} & 95.88\\
    $Camera 2$ & Res18; img128; hid32 & 95.44 & 92.74\\
    $Camera 3$ & Res18; img128; hid32 & 91.72 & 56.69\\
    \hline
    $All Cameras$ & Res18; img128; hid64 & 94.99 & \textbf{\underline{95.93}}\\
    $Camera 1$ & Res18; img128; hid64 & \textbf{95.61} & 91.26\\
    $Camera 2$ & Res18; img128; hid64 & 94.95 & 89.52\\
    $Camera 3$ & Res18; img128; hid64 & 93.02 & 83.57\\
    
    \bottomrule
    \end{tabular}
    \caption{\textbf{Metrics Table.} Quantitative analysis of both supervised and self-supervised classifiers from single and multiple cameras.}
    \label{tab:full_table}
\end{table}

\begin{table}[!htbp]
    \centering
    \begin{tabular}{c|c|c|c|c}
    \toprule
          & \multicolumn{2}{c}{\textbf{Supervised}} & \multicolumn{2}{c}{\textbf{Self-Supervised}} \\
    \midrule
    \textit{Dataset}  & \textit{Mean($\%$)} & \textit{CI(p=0.95)($\%$)} & \textit{Mean($\%$)} & \textit{CI(p=0.95)($\%$)} \\ \hline
    $All Cameras$  & $99.01\pm1.09$ & [98.18, 99.83] & $91.70\pm5.84$ & [85.87, 97.53]\\
    $Camera 1@Hid32$  & $99.76\pm0.28$ & [99.55, 99.97] & $82.70\pm8.60$ &  [74.11, 91.29]\\
    $Camera 2@Hid32$  & $98.72\pm0.95$ & [98.00, 99.43] & $84.36\pm6.80$ &  [77.57, 91.15]\\
    $Camera 3@Hid32$  & $98.45\pm1.39$ & [97.31, 99.58] & $52.95\pm13.37$ & [39.60, 66.31]\\
    \hline
    $All Cameras$ & $99.63\pm0.22$ & [99.47, 99.81] & $87.87\pm2.84$ & [85.03, 90.71]\\
    $Camera 1@Hid64$ & $99.91\pm0.08$ & [99.85, 99.98] & $76.34\pm9.60$ & [66.75, 85.92]\\
    $Camera 2@Hid64$  & $99.50\pm0.32$ & [99.26, 99.74] & $83.37\pm4.61$ &  [78.77, 87.98]\\
    $Camera 3@Hid64$  & $99.32\pm0.27$ & [99.12, 99.53] & $47.61\pm10.66$ & [36.96, 58.26]\\
    
    \bottomrule
    \end{tabular}
    \caption{\textbf{K-Fold Cross Validation Table.} K-Fold Cross Validation of both fully- and self-supervised frameworks. Shown are two hyper-parameter settings with Hid32 (upper rows) and Hid64 (lower rows). Subset Mean~(Mean) with Standard Deviation and Confidence Intervals~(CI) from single and multiple cameras.}
    \label{tab:kfcv_table}
\end{table}

    

\section{Discussions}
\label{sec:discussions}
\textbf{Reliable Automation of Training from Tracklets.} The findings from the self-supervision framework indicate that automated identification can be reliably achieved when presented with cattle walking between successive cameras at periodic intervals provided clean tracklets are available. This observation underscores the substantial potential for autonomous surveillance systems across whole farms having multi-camera coverage tailored to dairy cattle. Furthermore, our self-supervised results show that human intervention as part of the monitoring process can widely be eliminated given high accuracies of the system outputs and no requirement for labelling of any individual cattle during system training.

\textbf{Performance Improvements via Camera Integration.} In addition, the self-supervised framework demonstrates superior performance using information from multiple cameras, outperforming their individual camera counterparts due to the enriched feature diversity for each label. Performance is improved compared to previous studies in similar settings~\cite{gao2021towards, gao2022label}. \textcolor{black}{From the baseline of our study, the same methodology can be potentially expanded towards camera systems of diverse heights, angles and perspectives.} Conversely, the collective performance of the supervised classifier over the amalgamated camera dataset, while improved compared to some individual camera views, does not consistently surpass the performance when data from all cameras are used. As mentioned previously, the successful self-supervised training of the framework eliminates, the most time-consuming part of human labelling -- that is labelling individuals -- altogether. This, together with the performance gains made by merging data from disparate cameras, improves the viability of the visual ID approach on a commercial farm. \textcolor{black}{Additionally, despite contributing a significantly larger number of images and introducing dataset imbalance, Camera 3 exhibited noticeably poorer performance compared to the combined views. This analysis demonstrates that quantity imbalance alone does not necessarily improve single-camera performance over combined views, as the diversity of image features plays a crucial role in identification accuracy.}

\textcolor{black}{\textbf{Cross Validation.} The results of K-Fold cross-validation on the supervised pipeline indicate an overall increase in mean performance and stability compared to our standard training workflow. This improvement is attributed to the higher proportion of training data in the dataset compared to that in the standard workflow. With $k$ set to 10, 90$\%$ of the total data was used for training, which is greater than the proportion obtained through the uniform sampling used in the supervised workflow. As described in Section \ref{sec:results} - {\it Results}, we implemented a customized K-Fold cross-validation approach to replicate our designed training process, where $k$ was set to 7 to account for the full permutation of selecting one day as the test set. Due to the inherent daily biases in our dataset, this cross-validation setup resulted in higher variance and slightly lower average performance compared to the supervised pipeline.}

\subsection{Future Works}
\textbf{Open-Set Implementation.} In target re-identification applications, 
one of the main issues is that of how to deal with new individuals that were
never part of the original training set. This happens on dairy farms fairly
frequently as heifers enter the milking herd.
Identifying possible new individuals is known as the {\it open-set} problem~\cite{8953906, ghaffari2023towards, wang2024ultra}, and solving
this is an open challenge. In such scenarios, manual re-labelling of these unknown entities is not desirable. One possible solution to this problem is to build a classifier
capable of deciding whether a cow is indeed `new' or not, as in \cite{wang2024open}.

\textbf{Towards Fully Automatic Tracking.} In the context of proposing automated cattle identification methods, our current pipeline operates within certain constraints imposed by the available data. Firstly, the proximity of the selected cameras facilitates seamless transitions of cattle within frames, minimising abrupt changes in the morphology of cattle appearance. However, extending the framework to environments with cameras positioned with a greater degree of variance (either in distance or viewing angle) may necessitate extended training periods and could result in diminished performance due to the broader range of features that must be accommodated.
Secondly, the observed cattle typically traverse the monitored area exactly once during each period. This very constraint allows us to use the contrastive NTXent loss function; once a cow has been seen once, it must be different to all other cows observed over that period, and hence must reside in a unique part of the embedding space. In more open environments, for instance, in feeding and drinking zones, this would not necessarily be the case, and the same loss function would no longer be so appropriate. 
We also rarely see cows standing together very closely in our data, allowing
us to detect and localise them accurately. Once again, elsewhere on the farm, this may not be so true.

\section{Conclusion}
\label{sec:conclusions}
We introduced MultiCamCows2024, the first dataset consisting of multi-day images of Holstein-Friesian cows covered by multiple distinct cameras in a working farm. We performed supervised and self-supervised experiments using both merged and single cameras alone to test if fusing camera views enhances cow re-identification. We found that within our closed-set, fusing data across cameras boosts the performance of the cattle re-identification task
over that of simply using each view independently (assuming the same level of data utilisation). 



\newpage
\appendix

\section{Dataset Metadata} 
\label{Dataset Metadata}

\begin{figure}[!htbp]
\begin{subfigure}{0.49\textwidth}
\centering
\captionsetup{justification=centering}
\includegraphics[width=\textwidth]{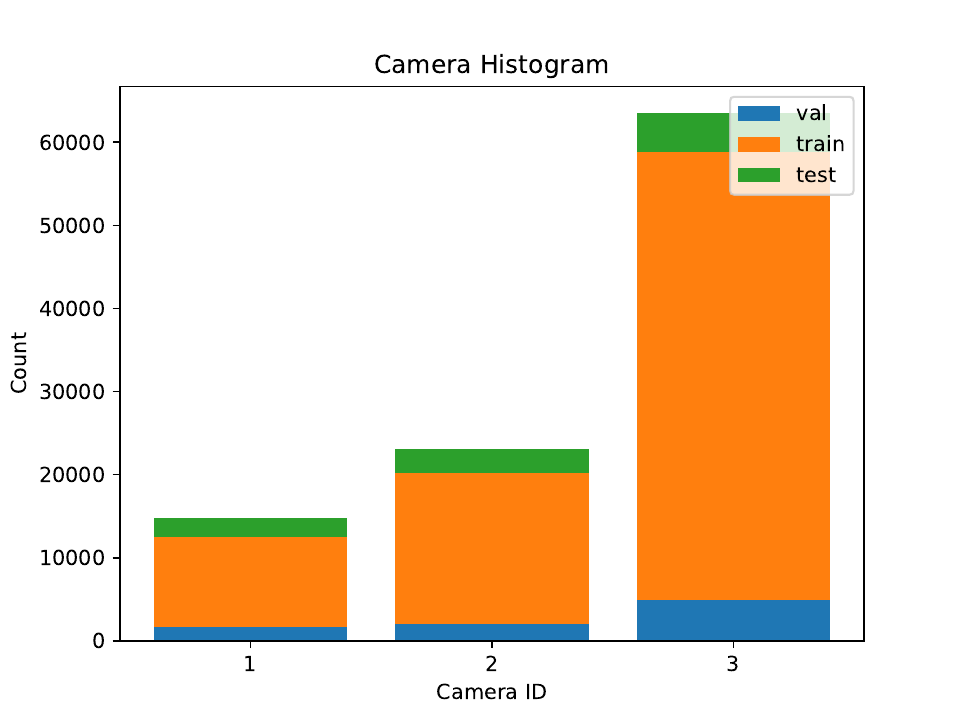}
\label{fig:camera_hist}
\end{subfigure}
\hfill
\begin{subfigure}{0.49\textwidth}
\centering
\captionsetup{justification=centering}
\includegraphics[width=\textwidth]{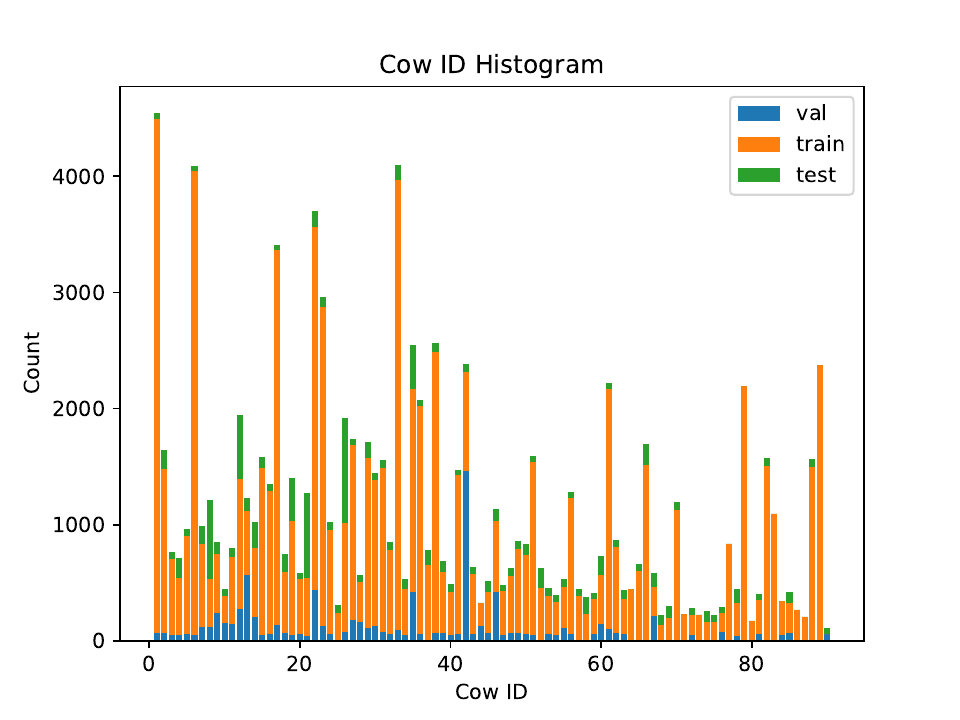}
\label{fig:id_hist}
\end{subfigure}
\caption{\textbf{Dataset Histogram across Cameras and Individuals.} Histogram showing the number of images in our dataset. (Left) Number of Images by camera and (Right) Number of images per ID. Much of the imbalance in numbers is caused by camera 3 which shows a part of the farm where cows are free to stop
walking after the milk-race - when this happens, given the fixed sampling rate, many more images are generated.}
\label{fig:metadata_hist}
\end{figure}

\begin{figure}[!htbp]
\centering
\captionsetup{justification=centering}
\begin{subfigure}{0.45\textwidth}
\centering
\includegraphics[width=0.8\textwidth]{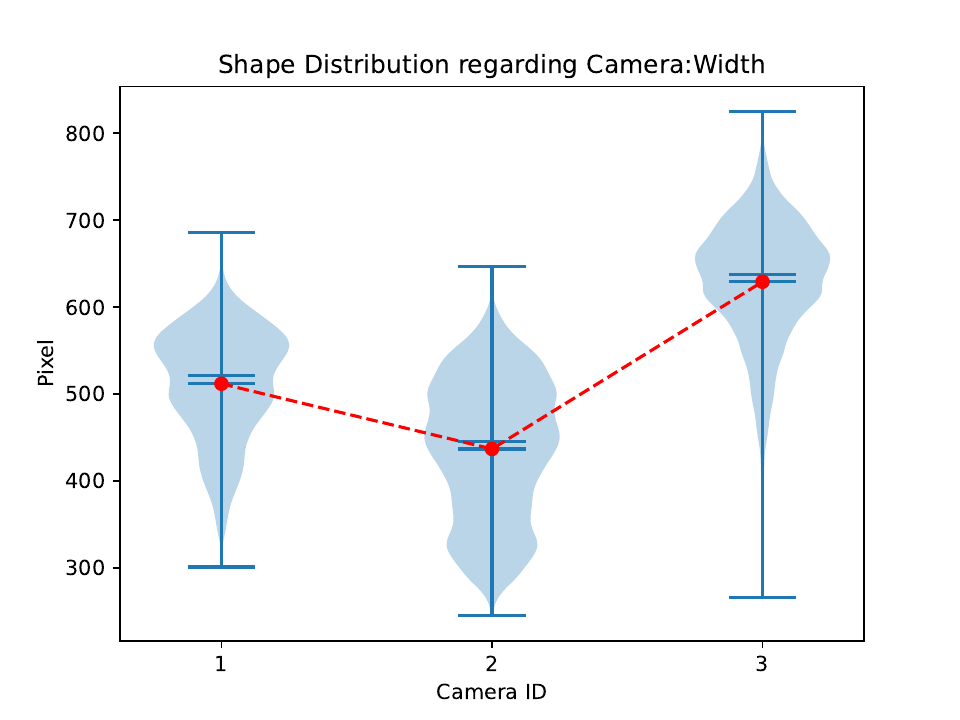}
\label{}
\end{subfigure}
\hfill
\begin{subfigure}{0.45\textwidth}
\centering
\includegraphics[width=0.8\textwidth]{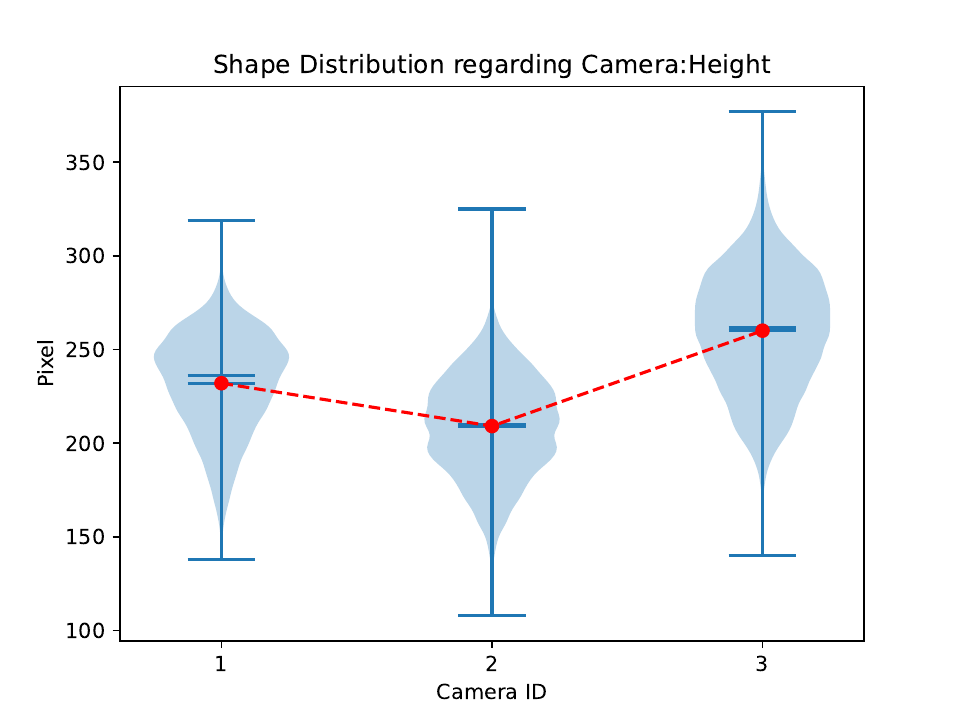}
\label{}
\end{subfigure}
\begin{subfigure}{0.45\textwidth}
\centering
\includegraphics[width=0.8\textwidth]{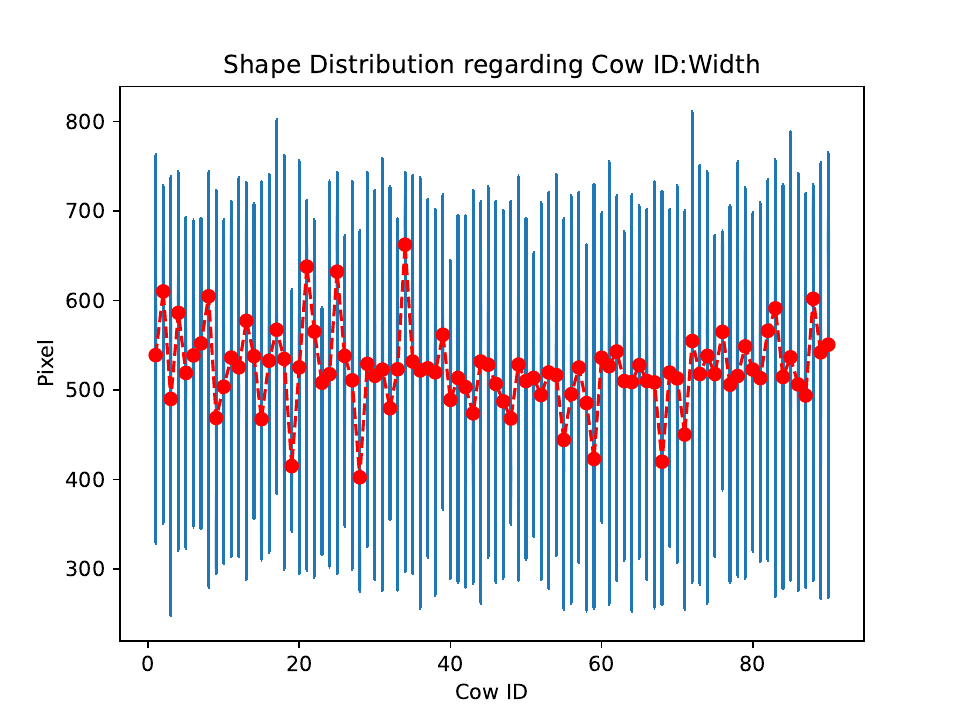}
\label{}
\end{subfigure}
\hfill
\begin{subfigure}{0.45\textwidth}
\centering
\includegraphics[width=0.8\textwidth]{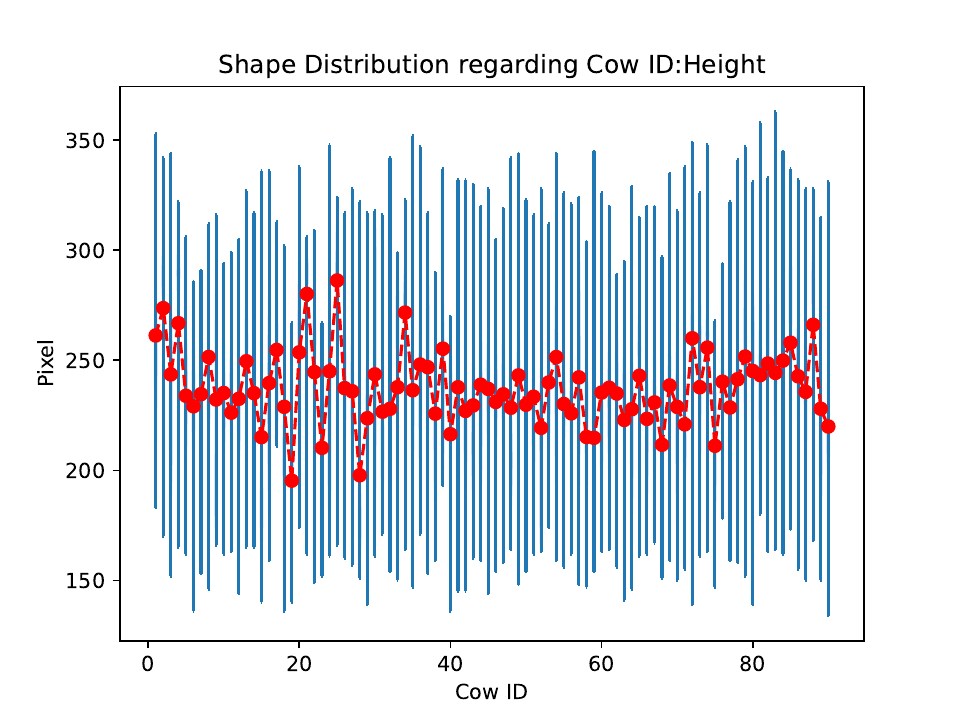}
\label{}
\end{subfigure}
\hfill
\begin{subfigure}{\textwidth}
\centering
\includegraphics[width=0.75\textwidth]{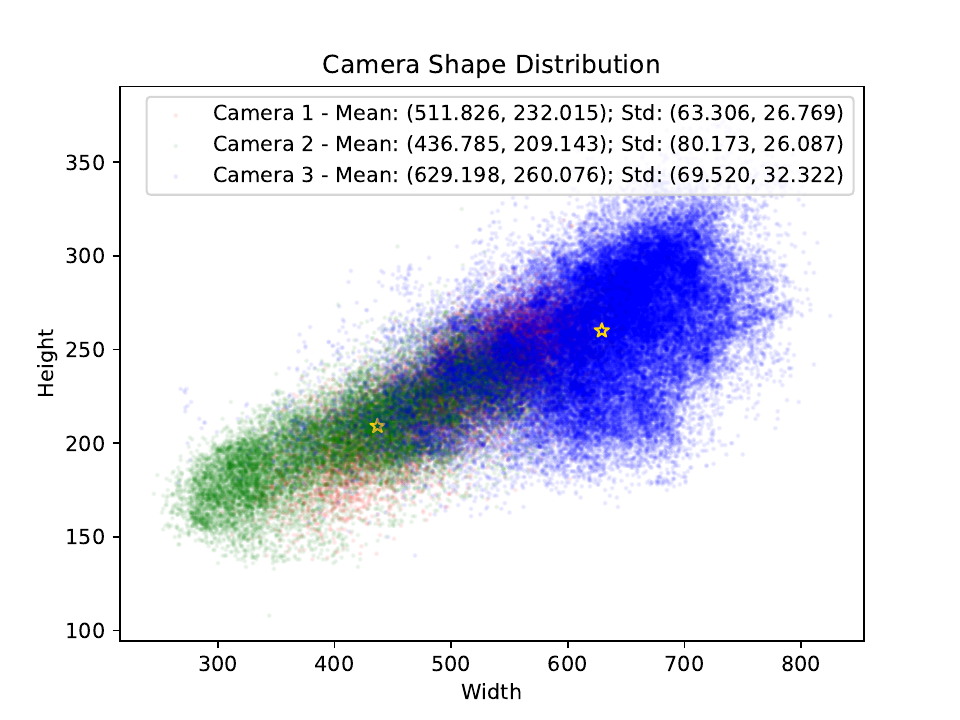}
\label{}
\end{subfigure}
\caption{\textbf{Dataset Metadata.} Metadata of width and height of cropped cow torsos from our dataset of $90$ unique cows. The upper row shows width and height information from different camera views. The middle and lower rows show shape distributions per ID.}
\label{fig:metadata_sizes}
\end{figure}
\newpage

\section{Identifier Embedding and Metrics Visualisation} \label{Visualisations}

\begin{figure}[!htbp]
\centering
\captionsetup{justification=centering}
\begin{tabular}{r p{\subfigexamplewidth} p{\subfigexamplewidth} p{\subfigexamplewidth} p{\subfigexamplewidth}}

Res18 & 
\begin{subfigure}[b]{\subfigexamplewidth}
    \caption{Fused}
    \vspace{5mm}
    \includegraphics[width=\subfigexamplewidth]{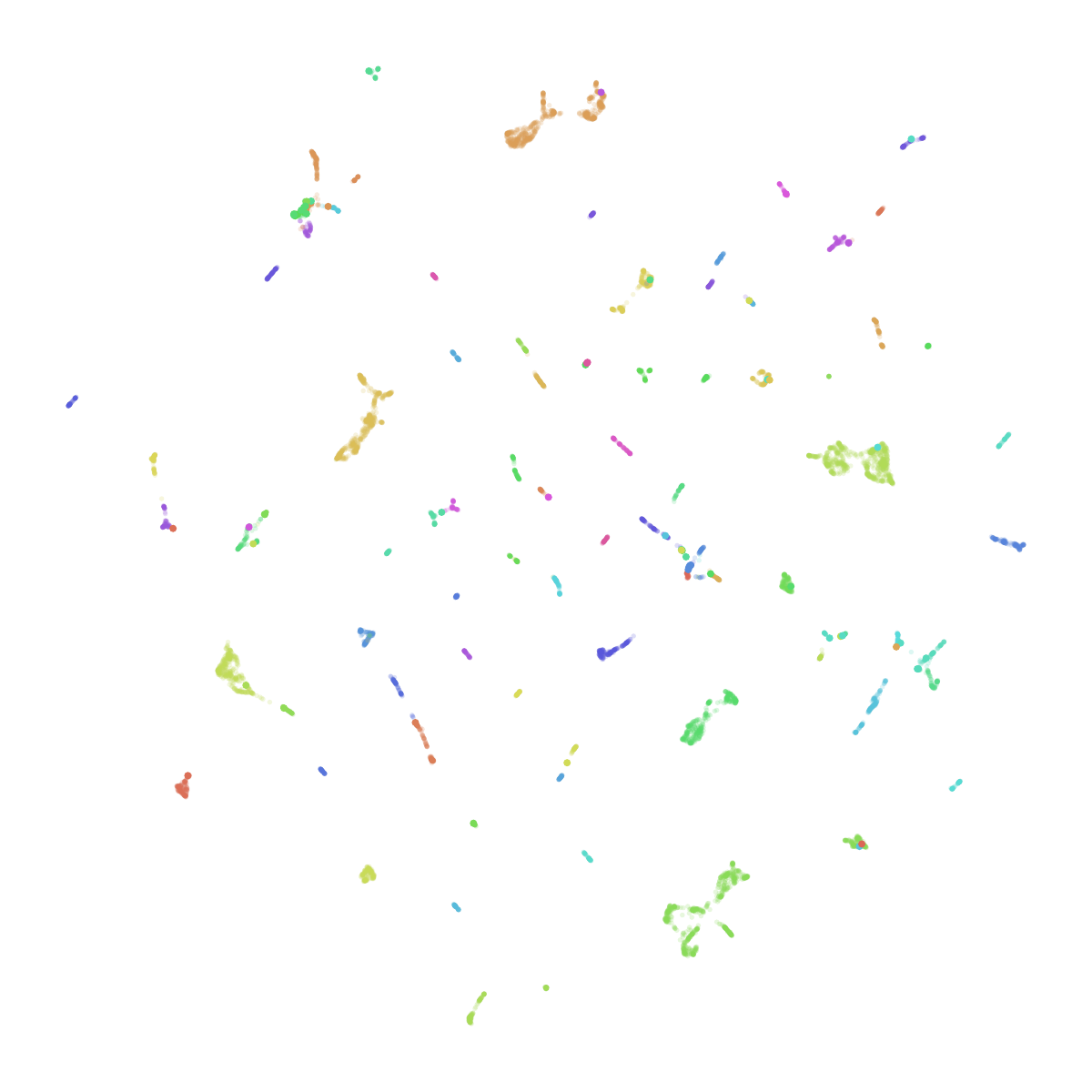}
\end{subfigure} &
\begin{subfigure}[b]{\subfigexamplewidth}
    \caption{Camera 1}
    \vspace{5mm}
    \includegraphics[width=\subfigexamplewidth]{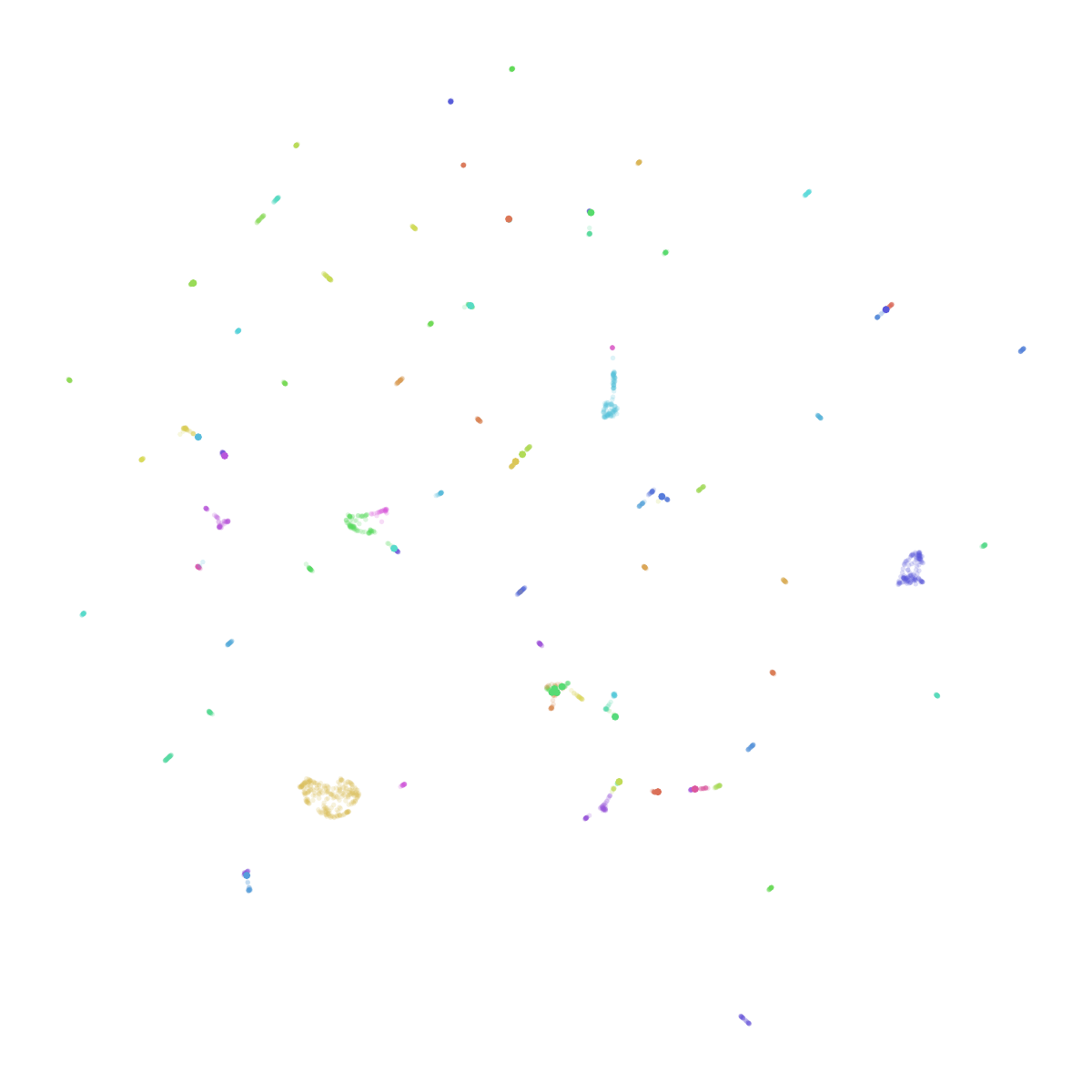}
\end{subfigure} &
\begin{subfigure}[b]{\subfigexamplewidth}
    \caption{Camera 2}
    \vspace{5mm}
    \includegraphics[width=\subfigexamplewidth]{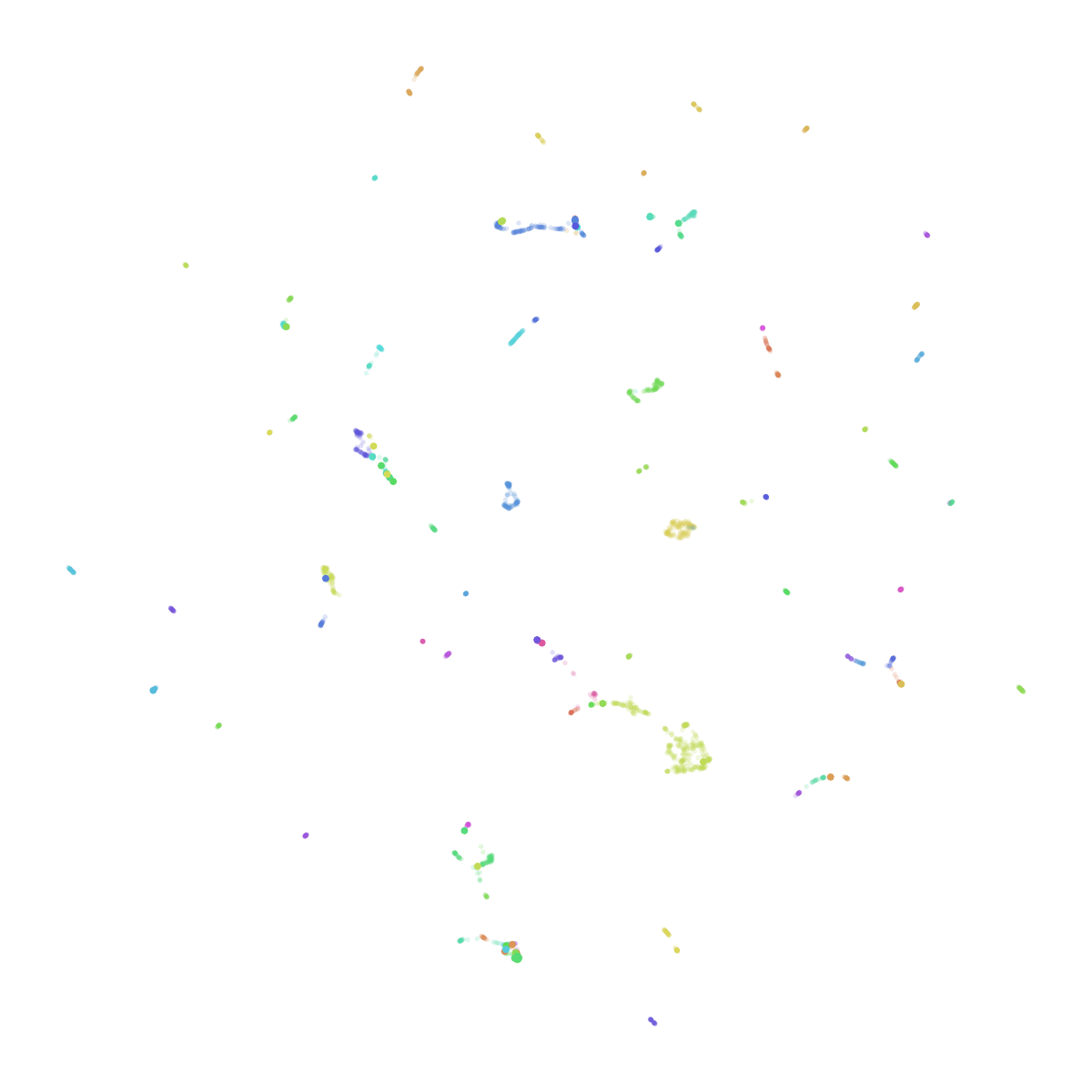}
\end{subfigure} &
\begin{subfigure}[b]{\subfigexamplewidth}
    \caption{Camera 3}
    \vspace{5mm}
    \includegraphics[width=\subfigexamplewidth]{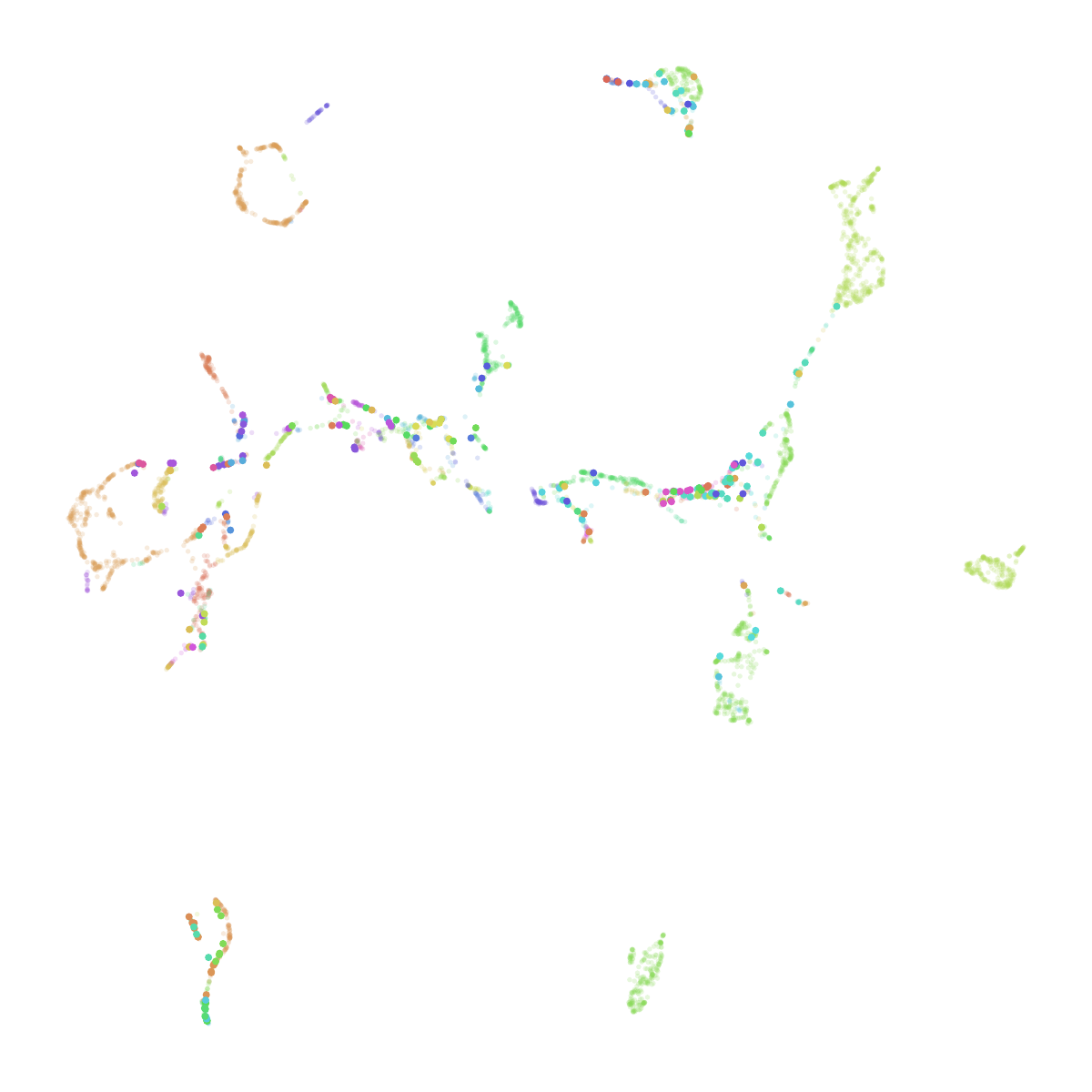}
\end{subfigure}
\\

Res50 & 
\includegraphics[width=\subfigexamplewidth]{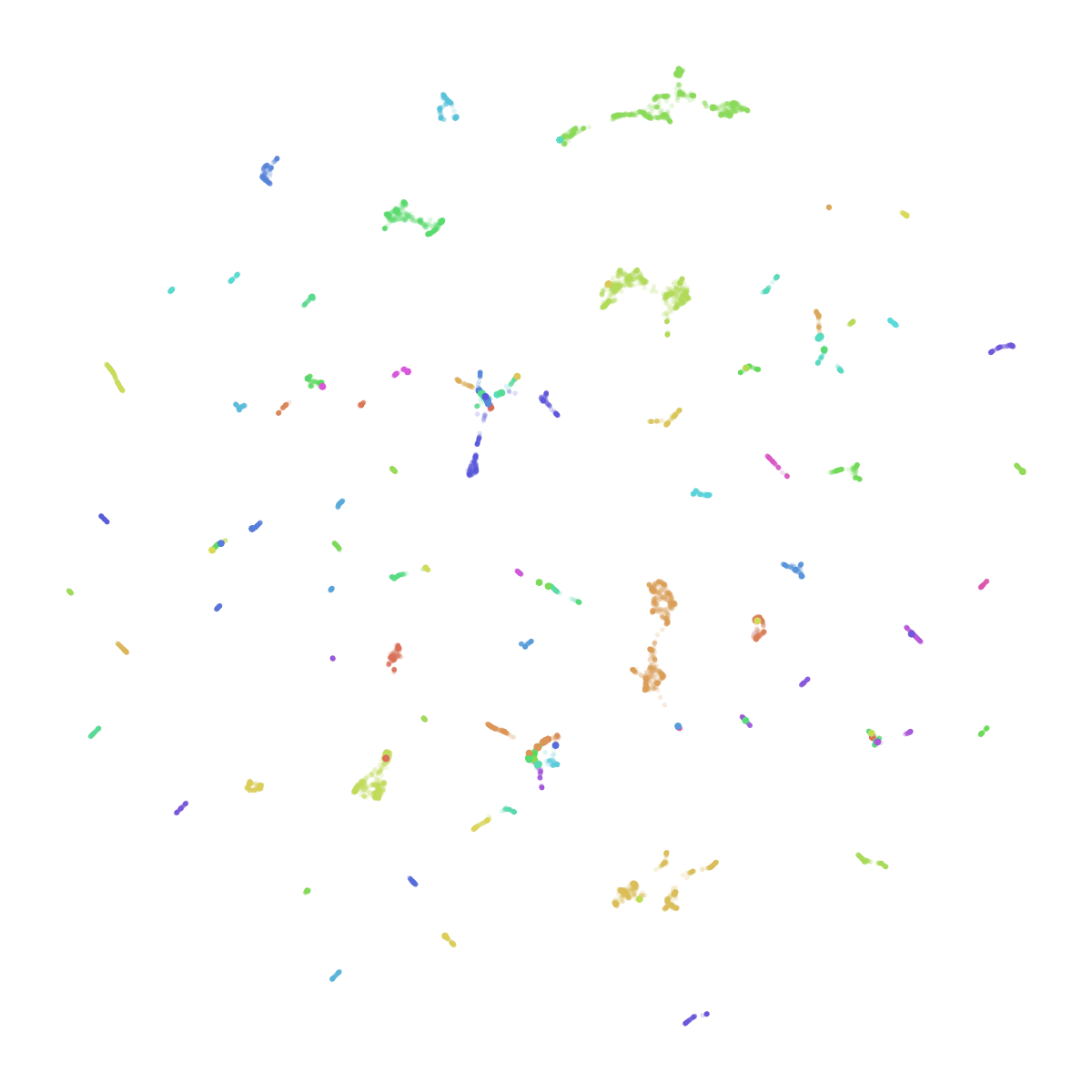} &
\includegraphics[width=\subfigexamplewidth]{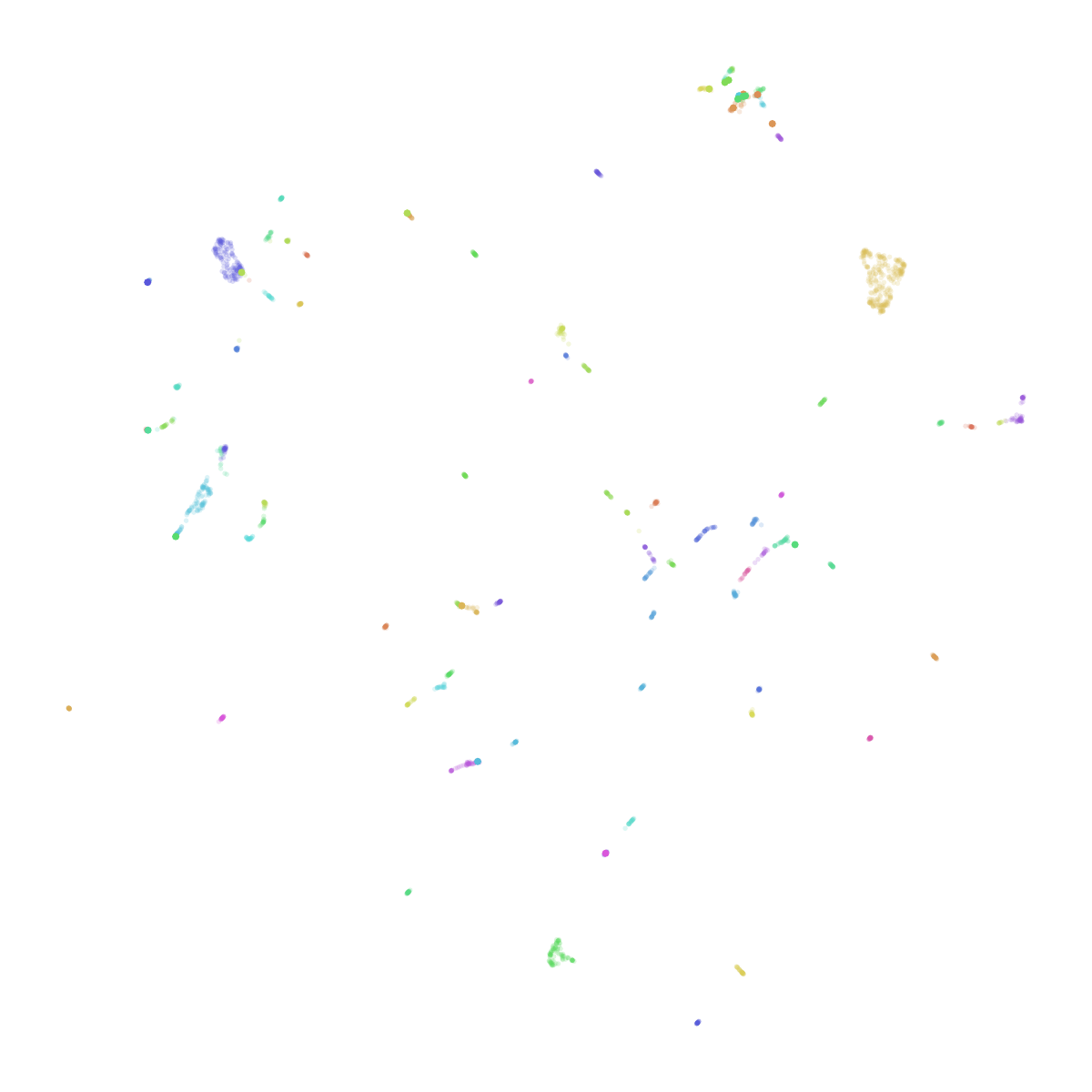} &
\includegraphics[width=\subfigexamplewidth]{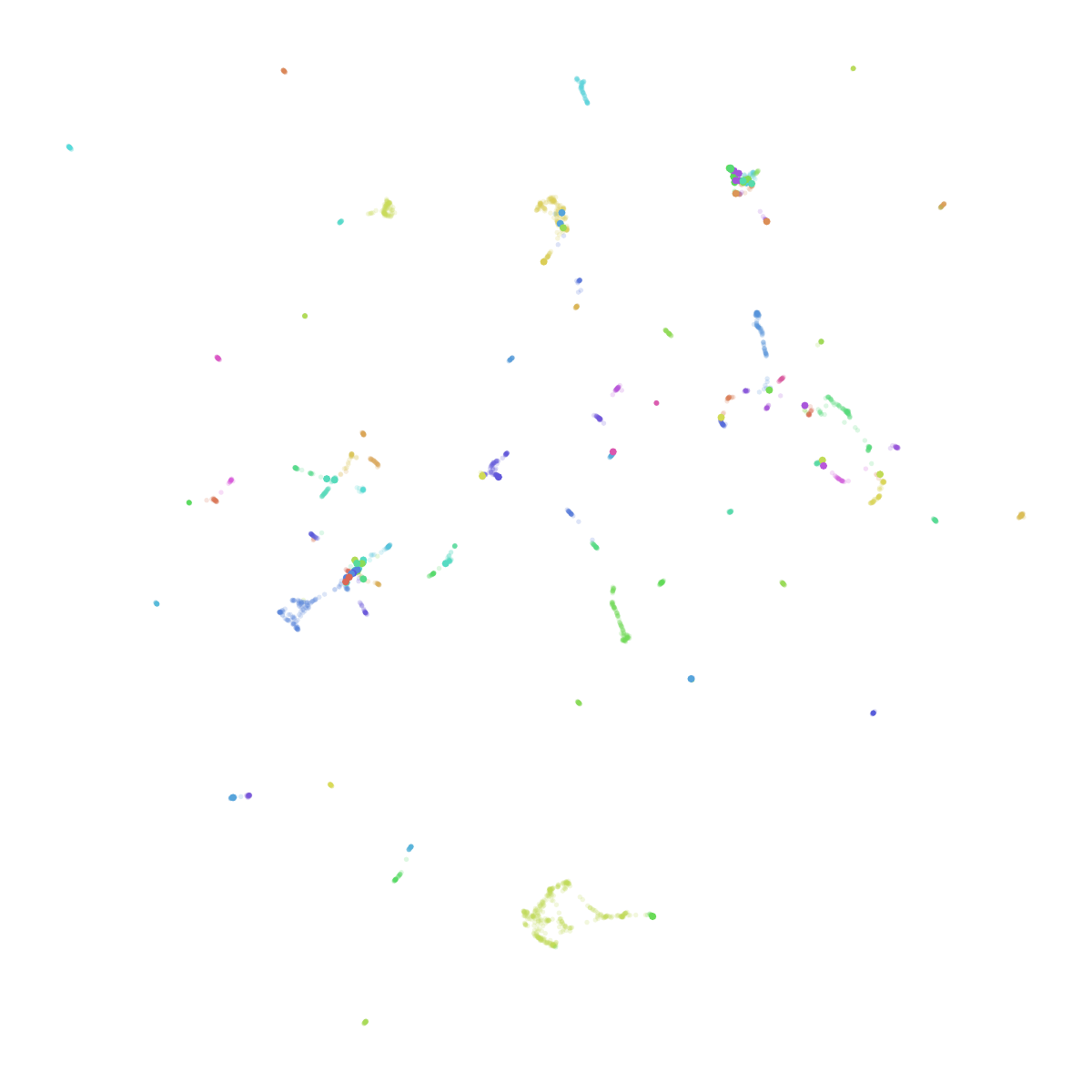} &
\includegraphics[width=\subfigexamplewidth]{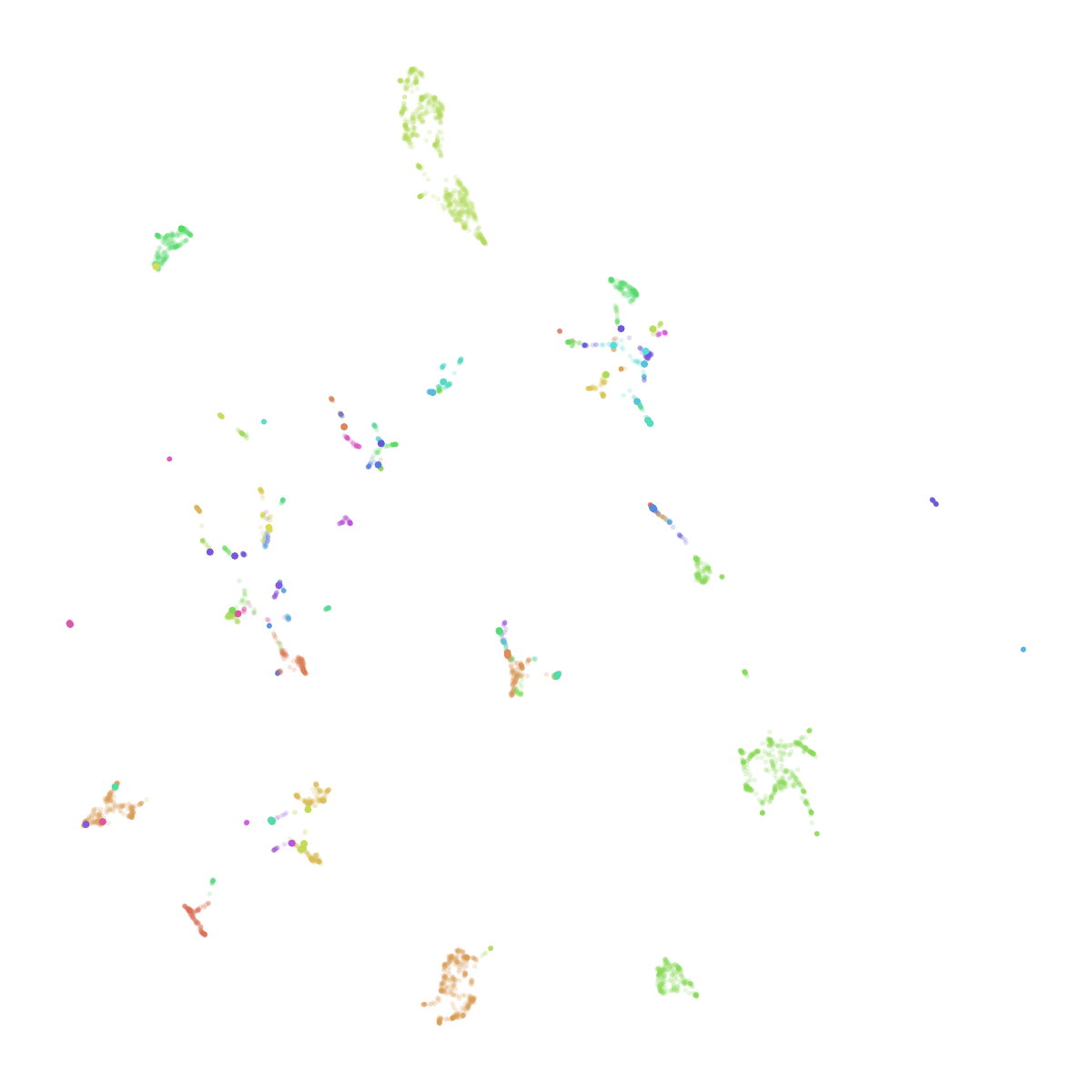}

\end{tabular}
\caption{\textbf{Visualisation of Embeddings.} Visualisation of embeddings for fused and individual camera perspectives. The false negatives are highlighted whilst dimming the true positives.}
\end{figure}

\clearpage
\bibliographystyle{elsarticle-num}
\bibliography{ref}

\newpage






\end{document}